\documentclass[pdflatex, sn-mathphys-ay]{sn-jnl} %
\usepackage{graphicx}%
\usepackage{multirow}%
\usepackage{amsmath,amssymb,amsfonts}%
\usepackage{amsthm}%
\usepackage{mathrsfs}%
\usepackage[title]{appendix}%
\usepackage{xcolor}%
\usepackage{textcomp}%
\usepackage{manyfoot}%
\usepackage{booktabs}%
\usepackage{algorithm}%
\usepackage{algorithmicx}%
\usepackage{algpseudocode}%
\usepackage{listings}%
\usepackage{subcaption, hyperref, microtype, arydshln} 
\usepackage{anyfontsize}

\theoremstyle{thmstyleone}%
\theoremstyle{thmstyletwo}%
\theoremstyle{thmstylethree}%
\DeclareMathAlphabet{\mathbbold}{U}{bbold}{m}{n}

\begin{document}

\author*[1]{\fnm{M. Ross} \sur{Kunz}}\email{matthew.kunz@inl.gov}
\author[1]{\fnm{John} \sur{Merickel}}\email{john.merickel@inl.gov}
\author[1]{\fnm{Keith} \sur{Wilson}}\email{keith.wilson@inl.gov}

\affil*[1]{\orgname{Idaho National Laboratory}} 

\title{Statistical Embeddings for Similarity, Retrieval, and Interpretable Alignment of Numeric Tabular Datasets}

\abstract{
    Numeric tabular datasets are the dominant data format in scientific practice, yet large language models lack native mechanisms for representing numeric datasets in a meaningful way across heterogeneous feature spaces. 
Existing approaches either target predictive modeling over individual datasets, which requires a shared set of variable definitions, or lack mechanisms for interpretable cross-dataset alignment. 
The proposed methodology characterizes numeric tabular datasets through structured exploratory data analysis descriptors, embeds those descriptors into a shared vector space using a pretrained sentence transformer, and quantifies cross-dataset similarity via Canonical Correlation Analysis (CCA). 
Furthermore, a penalized formulation of CCA is applied to recover sparse, interpretable variable-level correspondences between datasets, identifying which statistical descriptors or variable-level quantities drive cross-dataset alignment without requiring shared variable names or feature conventions.
Differential privacy is optionally applied to the descriptor set prior to embedding, supporting deployment in sensitive data contexts without requiring access to raw observations at time of comparison. 
The methodology is evaluated across 15 datasets spanning general-purpose benchmarks, materials informatics, and nuclear-grade graphite characterization. 
Results demonstrate a total $P@1$ score of 0.9, with known nearest-neighbor retrieval and cluster structure remaining robust across embedding ablations and differential privacy budgets.
The proposed framework provides a principled pathway for integrating heterogeneous numeric data into retrieval-augmented generation pipelines while preserving statistical context, with direct applications to data-driven algorithm selection and simulation model initialization for unknown datasets.

}


\maketitle

\section{Introduction}\label{sec:introduction}
Numeric tabular datasets are the dominant data format in scientific practice, yet organizing, searching, and comparing such datasets across heterogeneous feature spaces remains a largely unsolved problem in data management and knowledge discovery. 
Scientific data repositories accumulate datasets that vary in sample size, variable composition, measurement scale, and domain context, making it difficult to determine which datasets are structurally related, which share distributional properties, or which are likely to respond similarly to a given modeling approach. 
Existing dataset search and profiling tools produce human-readable summaries or rely on schema matching over shared variable names, neither of which supports similarity comparison across datasets with disjoint feature spaces. 
This limitation is especially pronounced in natural science domains, where a single physical system may be characterized through multiple experimental modalities with incompatible variable definitions, and where the cost of redundant or mismatched modeling efforts is high. 
A principled representation of dataset-level statistical structure comparable across heterogeneous feature spaces, and interpretable in terms of the underlying data properties, would substantially advance the ability to organize, retrieve, and reason over large collections of scientific tabular data.

Large language model (LLM) embedding spaces offer a potential pathway toward such a representation. 
By encoding text into vectors of common dimension, pretrained sentence transformers enable similarity comparisons across documents from disparate domains without requiring shared vocabulary or structure. 
This capability motivates a statistical embedding strategy: rather than representing a dataset by its raw observations or variable names, one characterizes it through structured exploratory data analysis (EDA) descriptors, serializes those descriptors into natural language sentences, and embeds the resulting sentences into the LLM's semantic space. 
The central challenge is ensuring that this embedding process preserves the statistical structure of the underlying data. 

The proposed methodology computes individual statistical descriptors for each numeric column alongside multivariate summary quantities, which together serve as the basis for embedding tabular data into a shared vector space. 
These embeddings support hierarchical clustering and Canonical Correlation Analysis (CCA) across datasets, analogous to multi-view document embeddings~\citep{dhillon2011multi}, even when data sources differ in dimension or variable composition. 
A penalized formulation of CCA is further employed to identify which variables or multivariate quantities drive cross-dataset correlation, yielding directly interpretable alignment structure. 
Differential privacy is optionally applied to the statistical descriptors prior to embedding, providing a principled mechanism for obfuscating sensitive summary information while preserving sufficient structure for meaningful cross-dataset comparison. 
Together, these components provide a means of interpretable tabular RAG that organizes heterogeneous numeric data while preserving statistical context.

The contributions of this work are as follows:
\begin{itemize}
    \item A structured EDA pipeline that produces statistically grounded, 
    sentence-serialized descriptors for numeric tabular datasets, with 
    optional differential privacy for sensitive data contexts, formalized 
    as a mapping from a raw data matrix to a collection of embedding 
    vectors in a shared semantic space.
    \item A cross-dataset similarity framework based on mean canonical 
    correlation between EDA embedding collections, supporting 
    nearest-neighbor retrieval and hierarchical clustering across datasets 
    with heterogeneous and disjoint feature spaces.
    \item A penalized CCA formulation yielding sparse canonical loadings 
    that identify which statistical descriptors or variable-level 
    quantities drive cross-dataset alignment, providing interpretability 
    not available under standard CCA.
    \item Empirical validation across 15 datasets spanning general-purpose 
    benchmarks, materials informatics, and nuclear-grade graphite 
    characterization, demonstrating a total $P@1$ score of 0.9 with nearest-neighbor retrieval.
    \item A framework for retrieval-based algorithm and simulation model 
    initialization, in which statistical similarity to cataloged datasets 
    with known modeling outcomes supports data-driven selection of 
    candidate methods for unknown datasets, connecting the proposed 
    approach to meta-learning and AutoML pipelines.
\end{itemize}

The remainder of this paper is organized as follows. Section~\ref{sec:related} 
reviews related work on tabular representation learning, dataset search, 
and differential privacy for data publishing. Section~\ref{sec:methodology} 
describes the EDA fingerprinting pipeline, embedding procedure, and 
penalized CCA similarity framework. Section~\ref{sec:datasources} details the 
datasets used for evaluation. Section~\ref{sec:results} presents 
nearest-neighbor retrieval results, hierarchical clustering analysis, and 
penalized CCA interpretability examples. Section~\ref{sec:conclusion} 
summarizes the findings and discusses directions for future work.

\section{Related Work}\label{sec:related}

\subsection{Tabular Representation and Foundation Models}
\label{subsec:related:tabular}

Several lines of research have addressed the challenge of representing 
tabular data within learned model architectures. TabNet~\citep{arik2021tabnet} 
introduced sequential attention over tabular features to enable 
instance-wise feature selection during training, improving interpretability 
for prediction tasks. SAINT~\citep{somepalli2021saint} 
extended the transformer architecture to tabular data through inter-sample 
and inter-feature attention, demonstrating competitive performance on 
supervised learning benchmarks. TabPFNv2~\citep{hollmann2025accurate} 
demonstrated that a transformer pretrained exclusively on synthetic tabular 
data achieves state-of-the-art predictive accuracy on small datasets via 
in-context learning. TabICL~\citep{qu2025tabicl} 
extended this paradigm to large datasets through a two-stage 
column-then-row attention mechanism. 
\cite{van2024tabular} argue that such Large Tabular Models (LTMs) 
remain significantly underrepresented in the foundation model literature 
despite the dominance of tabular data across scientific domains. 
Each of these approaches targets predictive modeling performance on individual datasets and does not address the problem of representing or comparing datasets across heterogeneous feature spaces.

\subsection{Dataset Search and Data Discovery}
\label{subsec:related:discovery}

The problem of organizing and retrieving datasets from large repositories 
has received sustained attention in the data management community. Dataset 
search over data lakes, collections of heterogeneous raw data assets 
stored without imposed schema, has been formalized as a problem of 
identifying semantically or structurally related tables from among 
thousands of candidates~\citep{nargesian2019data}. A central challenge in 
this setting is that related datasets often share neither variable names 
nor schema structure, requiring similarity measures that operate over 
content rather than metadata alone. Schema matching approaches address 
this by identifying correspondences between column names or value 
distributions across pairs of tables~\citep{rahm2001survey}, but these 
methods rely on lexical overlap or value-level comparison and do not 
generalize to datasets with entirely disjoint vocabularies. Dataset join 
and unionability discovery methods~\citep{zhu2019josie,nargesian2019data} 
extend schema matching to identify tables that can be joined or stacked, 
but again require overlapping values or column semantics and are not 
designed for cross-domain similarity assessment. 
The ARDA system~\citep{chepurko2020arda} and related 
augmentation frameworks treat dataset discovery as a feature engineering 
problem, identifying external tables that improve downstream model 
performance when joined to a query table, but this framing assumes a 
fixed target task rather than general-purpose similarity assessment. 
More recent work on dataset discovery for machine learning~\citep{zha2025data} 
has begun to treat datasets as first-class objects with learnable 
representations, but existing approaches remain largely confined to 
relational or structured tabular settings with compatible schemas. 
Existing profiling approaches produce human-readable summaries rather 
than embedding-compatible representations suitable for retrieval over 
heterogeneous collections.

\subsection{Tabular Retrieval-Augmented Generation and Embedding}
\label{subsec:related:rag}

Retrieval-augmented generation (RAG) offers a scalable approach to 
incorporating external knowledge at inference time by retrieving relevant 
context and injecting it into the model's input~\citep{lewis2020retrieval}. 
Applied to tabular data, RAG pipelines retrieve relevant rows, columns, 
or dataset chunks at query time, but standard embedding models are 
pretrained predominantly on text corpora and underperform on numeric or 
relational table content. The Tabular Embedding Model 
(TEM)~\citep{khanna2025tabular} addresses this by fine-tuning lightweight 
embedding models specifically for tabular RAG pipelines, achieving 
substantial retrieval gains over general-purpose embedders. 
However, TEM operates at the row level within a single dataset 
and does not address cross-dataset similarity. 
The vision-language model connector 
paradigm~\citep{zhang2024vision,wu2019unified} demonstrates that 
heterogeneous modalities can be aligned into a shared latent space 
through learned projection modules, but \cite{li2025lost} show 
that such projections distort local embedding geometry substantially. 
This geometric distortion finding motivates the use of 
distributional statistics rather than raw numeric values as the 
embedding substrate, as statistical descriptors are more naturally 
expressible in the semantic space of a pretrained sentence transformer 
than are raw numeric observations. No existing tabular RAG framework 
produces dataset-level embeddings that support cross-dataset similarity 
comparison while preserving interpretable statistical structure.

\section{Methodology}\label{sec:methodology}

The proposed methodology is organized into two primary components. The first concerns the 
statistical characterization of numeric tabular datasets through a structured exploratory 
data analysis (EDA) framework. The second component addresses 
the transformation of these descriptors into a common vector space suitable for cross-dataset 
comparison and retrieval. Pairwise dataset similarity is then quantified 
through Canonical Correlation Analysis (CCA) with a sparse low dimensional approximation for interpretability.

\subsection{Problem Formulation}
\label{subsec:method:problem}

Let $\mathcal{D} = \{X \in \mathbb{R}^{n \times p}\}$ denote a numeric 
tabular dataset with $n$ observations and $p$ variables. 
The goal is to 
construct a mapping $\phi: \mathcal{D} \rightarrow \mathbb{R}^{d}$ such 
that for a collection of datasets $\{\mathcal{D}_1, \ldots, \mathcal{D}_N\}$ 
with potentially disjoint variable sets, the pairwise similarity 
$s(\mathcal{D}_i, \mathcal{D}_j) = f\!\left(\phi(\mathcal{D}_i), 
\phi(\mathcal{D}_j)\right)$ reflects meaningful statistical relationships 
between datasets even when $p_i \neq p_j$ or the variable sets are 
entirely disjoint. 
The mapping $\phi$ must satisfy three properties. 
First, it must be computable without access to raw observations at 
comparison time, supporting privacy-preserving retrieval over sensitive 
data collections. Second, the resulting embedding must be compatible with 
a pretrained sentence transformer's semantic space, enabling integration 
with downstream LLM-based retrieval pipelines. Third, the similarity 
measure $s$ must yield interpretable alignment structure that is 
recoverable via sparse canonical analysis, identifying which statistical 
descriptors or variable-level quantities drive cross-dataset correlation. 
Sections~\ref{sec:methodology:eda} and~\ref{sec:methodology:cca} describe 
the construction of $\phi$ and $s$ respectively.

A common theme throughout the methodology is the use of penalized 
regression and matrix decomposition techniques, chosen primarily for 
 interpretability. The 
unifying computational primitive is the singular value decomposition 
(SVD), which factorizes a data matrix $X \in \mathbb{R}^{n \times p}$ 
as:
\begin{equation}
    X = U \Sigma V^\top,
\end{equation}
where $U \in \mathbb{R}^{n \times r}$ contains the left singular 
vectors, $\Sigma \in \mathbb{R}^{r \times r}$ is a diagonal matrix of 
ordered singular values $\sigma_1 \geq \sigma_2 \geq \cdots \geq 
\sigma_r \geq 0$, $V \in \mathbb{R}^{p \times r}$ contains the right 
singular vectors, and $r = \min(n, p)$ is the rank of $X$. Each 
application of the SVD in the proposed methodology serves a distinct 
interpretive purpose in different stages, e.g., imputation or cross-correlation between datasets. 
The general penalized regression problem that unifies these applications takes the form:
\begin{equation}
    \min_{\beta} \; \mathcal{L}(X, \beta) + 
    \sum_{j} p_{\lambda}(|\beta_j|),
\end{equation}
where $\mathcal{L}(X, \beta)$ is a loss function measuring fit to the 
data, $\beta$ is the parameter vector of interest, $p_{\lambda}$ is a 
penalty function controlled by regularization parameter $\lambda > 0$, 
and the sum is taken over the components of $\beta$. The choice of 
$\mathcal{L}$ and $p_{\lambda}$ can take on different forms depending on the
application in the next subsections, e.g., $\ell_1$ regularization for 
enforcement of sparsity. Together, these 
instantiations ensure that interpretability is preserved at every stage 
of the pipeline, from imputation through descriptor computation to 
cross-dataset alignment.

\subsection{Statistical Fingerprinting via EDA}\label{sec:methodology:eda}

Prior to dataset analysis, a lightweight structural assessment is 
performed to characterize the data matrix and guide downstream 
computation. Let $X \in \mathbb{R}^{n \times p}$ denote a numeric 
tabular dataset with $n$ observations and $p$ variables. Missing values 
are tabulated across columns and imputed via soft-thresholding on the 
singular value decomposition~\citep{mazumder2010spectral} before any 
further computation is performed, as missingness in the raw matrix would 
otherwise corrupt the singular value spectrum used for rank estimation 
and multivariate characterization. Specifically, the imputed matrix 
$\tilde{X} \in \mathbb{R}^{n \times p}$ is obtained by applying a 
soft-threshold operator to the singular values of the incomplete matrix, 
replacing missing entries with values consistent with a low-rank 
approximation. The numerical rank of $\tilde{X}$ is then estimated 
using the Gavish--Donoho optimal hard threshold~\citep{gavish2014optimal}. 
Letting $\sigma_{\text{med}}$ denote the median of the observed singular 
values of $\tilde{X}$, the 
threshold $\hat{\tau}$ is given by:
\begin{equation}
    \hat{\tau} \approx 2.858 \cdot \sigma_{\text{med}},
\end{equation}
 and the numerical rank $\hat{r}$ is estimated as:
\begin{equation}
    \hat{r} = \sum_{k} \mathbf{1}[\sigma_k > \hat{\tau}].
\end{equation}
Differential privacy is optionally applied at this stage to bound the 
sensitivity of subsequent summary statistics, ensuring that aggregate 
descriptors do not expose individual 
observations~\citep{dwork2025differential}. For each scalar descriptor 
$d_k$ with global sensitivity $\Delta_k$, where $\Delta_k$ denotes the 
maximum change in $d_k$ due to the addition or removal of a single 
observation, Laplace noise calibrated to $\Delta_k / \epsilon$ is added 
prior to sentence serialization. Here $\epsilon > 0$ controls the 
privacy budget, with smaller values providing stronger privacy 
guarantees at the cost of greater descriptor perturbation.

Multivariate analysis is then performed to characterize the global
structure and variable relationships within $\tilde{X}$. Matrix-level
norms and spectral quantities of $\{\sigma_k\}$ are computed to
summarize the overall scale, energy, and dominant directional structure;
for example, the spectral norm is defined as
$\|\tilde{X}\|_2 = \sigma_1$, the largest singular value. A full list
of multivariate measures is provided in the first column of
Table~\ref{tab:munge_descriptors}. Prior to multivariate analysis,
pairwise Pearson correlations are evaluated across all numeric columns,
and variables exceeding a collinearity threshold (e.g., $R^2 > 0.95$)
are removed to avoid redundancy and suppress dominant pairwise structure
that would otherwise obscure smaller multivariate interactions. This
yields a reduced column set $\mathcal{N}^* \subseteq \{1, \ldots, p\}$.
Multivariate variable importance is subsequently estimated using
smoothly clipped absolute deviation (SCAD) penalized
regression~\citep{fan2001variable, breheny2011coordinate}, with $a =
3.7$ and penalty parameter $\lambda$ selected via cross-validation.
SCAD is preferred over $\ell_1$-based alternatives such as
LASSO~\citep{tibshirani1996regression} because its non-concave form
reduces estimation bias and produces sparser solutions through the
oracle property~\citep{fan2001variable}. SCAD is not a strict requirement and the variable selection
technique may be substituted depending on the domain, provided the
objective remains identification of characteristic variable correlation
fingerprints.

Univariate statistical descriptors are computed for each numeric 
variable to characterize its marginal distribution and temporal or 
sequential behavior. This includes standard exploratory quantities such 
as bounds, central moments, vector norms, and quantile summaries, as 
well as information-theoretic measures including entropy; a full list 
is provided in the second column of Table~\ref{tab:munge_descriptors}. 
Robust statistics, e.g., median and median absolute deviation, are 
computed alongside their classical counterparts to provide 
outlier-resistant summaries of location and spread. For variables with 
sequential structure, autocorrelation coefficients are estimated to 
quantify serial dependence, and the Fast Fourier Transform (FFT) is 
applied to identify dominant periodic components and characterize the 
frequency content of each column. Change point detection is applied to 
identify structural breaks in the mean, variance, and joint 
mean-variance behavior of each column, yielding segment-level summaries 
that capture non-stationarity~\citep{killick2012optimal}. Specifically, 
the PELT algorithm with a BIC penalty is used to detect mean, variance, 
and mean-variance shifts, implemented via the \texttt{changepoint} 
package in R~\citep{killick2014changepoint}.

Categorical variables, identified through cardinality thresholding at 
$\kappa$, receive dedicated treatment that combines column-level 
discrete summaries with partition-level continuous analysis. Let 
$\mathcal{C} = \{j : |\text{unique}(X_{:,j})| \leq \kappa\}$ denote 
the set of categorical column indices and 
$\mathcal{N} = \{1,\ldots,p\} \setminus \mathcal{C}$ the numeric 
columns. For each categorical column $j \in \mathcal{C}$, column-level 
descriptors include the absolute and relative frequency distribution 
over observed levels, the modal category and its frequency, and the 
number of unique levels. Additionally, for each category level $c$ with 
at least 30 observations, the full univariate and multivariate 
descriptor sets described above are recomputed within the subset 
$\{X_{i,:} : X_{i,j} = c\}$, providing distributional characterization 
of each sufficiently represented group. Category levels with fewer than 
30 observations are excluded from partition-level analysis to ensure 
descriptor stability, as moment and quantile estimates are unreliable 
at small sample sizes. For datasets containing a class variable, the 
class column is treated as categorical regardless of its numeric 
encoding, and partition-level descriptors are computed for each class 
level meeting the sample size threshold.

Data segments identified through change point detection receive 
analogous treatment. For each contiguous segment $s$ identified within 
a numeric column, both the multivariate and univariate descriptor sets 
are recomputed within the segment, allowing the method to represent 
local distributional structure that would otherwise be obscured by 
global summaries computed over the full variable range. Segment-level 
descriptors are serialized using the same natural language templates as 
the global descriptors, with the segment index appended to the variable 
name token to distinguish local from global summaries.

Each statistical descriptor produced by the analysis pipeline is paired 
with a natural language sentence that contextualizes the numeric 
quantity. These sentences follow structured templates calibrated to each 
measure; for example, a singular value threshold entry might read:
\begin{quote}
The optimal singular value threshold is 23.4378; singular values below 
this are considered noise under the Gavish--Donoho criterion.
\end{quote}
while a distributional descriptor might read:
\begin{quote}
Kurtosis is 2.4103; the distribution is platykurtic with lighter tails 
than normal.
\end{quote}
The variable name or column header is then prepended to the sentence 
to supply feature-level context. For example:
\begin{quote}
Variable: pressure. Measure: kurtosis. Response: Kurtosis is 2.4103; 
the distribution is platykurtic with lighter tails than normal.
\end{quote}
For multivariate or matrix-level descriptors, where no single variable 
can be attributed, the variable name is replaced with the token 
\textit{matrix} to distinguish aggregate quantities from column-specific 
ones. Let $\mathcal{S} = \{t_1, t_2, \ldots, t_M\}$ denote the 
resulting ordered collection of descriptor sentences, where $M$ denotes 
the total number of sentences produced by the pipeline and depends on 
$p$, $n$, the number of detected change point segments, and the number 
of categorical levels meeting the sample size threshold. The embedding 
of $\mathcal{S}$ into a shared vector space is described in 
Section~\ref{sec:methodology:cca}.

\begin{figure}[htp]
    \centering
    \includegraphics[width=0.95\linewidth]{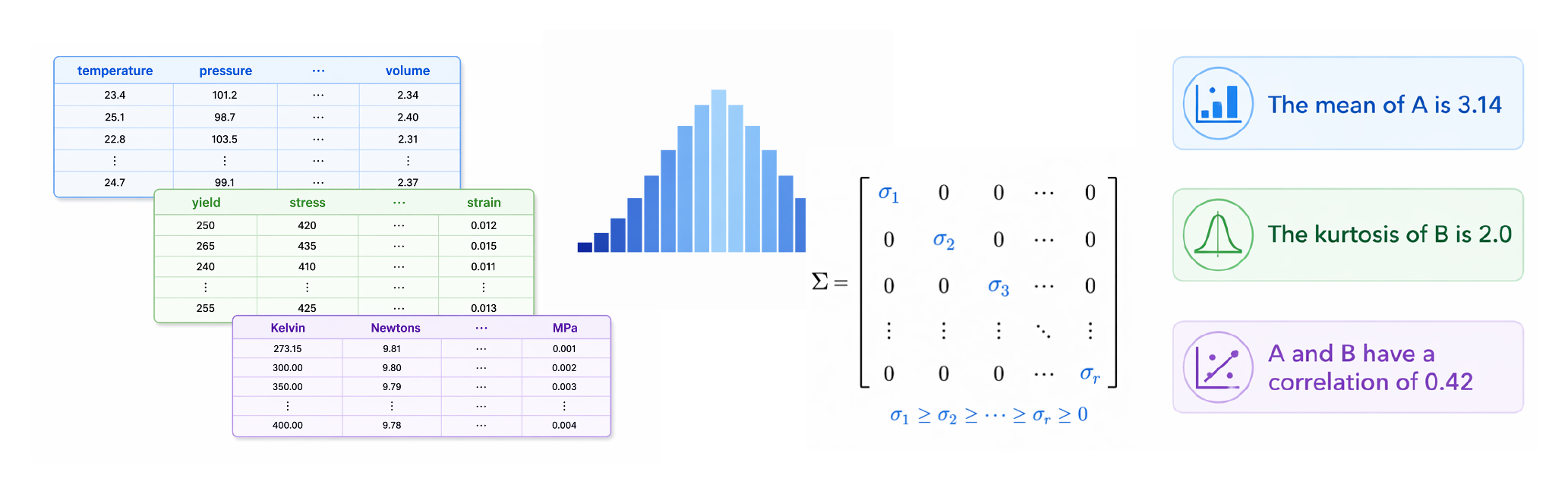}
    \caption{Statistical characterization process flow for moving from 
             raw data to embeddings.}
    \label{fig:statistical_characterization}
\end{figure}

\begin{algorithm}[H]
\caption{EDA Fingerprinting and Sentence Serialization}
\label{alg:eda_embedding}
\begin{algorithmic}[1]

\Require Data matrix $X \in \mathbb{R}^{n \times p}$; privacy flag 
         $\texttt{dp} \in \{\texttt{true}, \texttt{false}\}$; 
         privacy budget $\epsilon > 0$ (used only if 
         $\texttt{dp} = \texttt{true}$); 
         correlation threshold $\rho \in (0, 1)$; 
         cardinality threshold $\kappa \in \mathbb{Z}^+$;
         minimum partition size $n_{\min} = 30$

\Ensure Ordered sentence collection 
        $\mathcal{S} = \{t_1, \ldots, t_M\}$, where 
        $M$ denotes the total number of descriptor sentences 
        produced by the pipeline

\vspace{0.5em}
\State Tabulate missing value counts per column; impute via 
       soft-thresholded SVD~\citep{mazumder2010spectral} 
       to obtain $\tilde{X}$, prior to all subsequent 
       computation

\State Compute SVD of imputed matrix $\tilde{X} \approx 
       U\Sigma V^\top$, where $U \in \mathbb{R}^{n \times r}$, 
       $\Sigma \in \mathbb{R}^{r \times r}$, and 
       $V \in \mathbb{R}^{p \times r}$; estimate numerical 
       rank $\hat{r}$ via the Gavish--Donoho 
       threshold~\citep{gavish2014optimal}

\State Identify categorical columns: 
       $\mathcal{C} = \{j : |\text{unique}(X_{:,j})| \leq \kappa\}$; 
       let $\mathcal{N} = \{1,\ldots,p\} \setminus \mathcal{C}$

\If{$\texttt{dp} = \texttt{true}$}
    \State Add Laplace noise $\text{Lap}(\Delta_k / \epsilon)$ 
           to each scalar descriptor $d_k$, where $\Delta_k$ 
           is the global sensitivity of $d_k$
\EndIf

\vspace{0.5em}
\State \textit{// Multivariate Descriptors}

\State Compute matrix-level norms and spectral quantities 
       for $\tilde{X}$ (see Table~\ref{tab:munge_descriptors})

\State Compute pairwise Pearson correlations among 
       $\mathcal{N}$; remove columns exceeding $\rho$ to 
       form $\mathcal{N}^*$; estimate multivariate variable 
       importance via SCAD~\citep{fan2001variable} over 
       $\tilde{X}^*$

\vspace{0.5em}
\State \textit{// Univariate Descriptors}

\For{each column $j \in \mathcal{N}$}
    \State Compute univariate descriptors for $\tilde{X}_{:,j}$ 
           (see Table~\ref{tab:munge_descriptors})
    \State Apply PELT with BIC penalty to $\tilde{X}_{:,j}$ 
           for mean, variance, and joint mean-variance 
           shifts~\citep{killick2014changepoint}; record 
           segment boundary locations
\EndFor

\vspace{0.5em}
\State \textit{// Categorical and Segment Descriptors}

\For{each categorical column $j \in \mathcal{C}$}
    \State Compute frequency distribution, modal category, 
           and unique level count for $X_{:,j}$
    \For{each level $c$ where 
         $|\{i : X_{i,j} = c\}| \geq n_{\min}$}
        \State Recompute univariate and multivariate 
               descriptors within 
               $\{\tilde{X}_{i,:} : X_{i,j} = c\}$
    \EndFor
\EndFor

\For{each segment $s$ identified by change point detection}
    \State Recompute univariate and multivariate descriptors 
           within segment $s$; append segment index to 
           variable name token
\EndFor

\vspace{0.5em}
\State \textit{// Sentence Serialization}

\State Initialize $\mathcal{S} \leftarrow [\,]$
\For{each descriptor $d_k$ with variable name or 
     token \texttt{matrix}}
    \State Instantiate natural language template for $d_k$; 
           produce sentence $t_k$; append to $\mathcal{S}$
\EndFor

\State \Return $\mathcal{S}$

\end{algorithmic}
\end{algorithm}

\vspace{0.5em}
\noindent The statistical descriptor computation described in 
Algorithm~\ref{alg:eda_embedding} is implemented in the \texttt{mungeR} 
package, an open-source R package providing modular utilities for 
numeric and general-purpose embedding of tabular datasets \citep{kunz2026mungeR}. The package 
includes all descriptor functions, sentence serialization templates, 
and differential privacy wrappers. 


\subsection{Embedding, Hierarchical Clustering, and CCA}\label{sec:methodology:cca}

Each descriptor sentence $t_k \in \mathcal{S}$ produced by the 
fingerprinting pipeline is encoded into a fixed-length embedding vector 
using a pretrained sentence transformer $\mathcal{E}: \mathcal{T} 
\rightarrow \mathbb{R}^{d_{\mathcal{E}}}$, where $\mathcal{T}$ denotes 
the space of text strings and $d_{\mathcal{E}}$ is the output dimension 
of the chosen model. In this study, the \texttt{all-MiniLM-L6-v2} 
model is used, which produces $d_{\mathcal{E}} = 384$-dimensional 
embeddings and is publicly available through Hugging 
Face~\citep{wang2020minilm}. This choice is not prescriptive; any 
sentence transformer producing embeddings of consistent dimension may 
be substituted depending on computational constraints or domain 
requirements. The embedding collection for dataset $\mathcal{D}_i$ is 
defined as:
\begin{equation}
    \Phi_i = [\mathbf{v}_1^{(i)}; \cdots; \mathbf{v}_{M_i}^{(i)} ] \subset 
    \mathbb{R}^{d_{\mathcal{E}}}, 
    \quad \mathbf{v}_k^{(i)} = \mathcal{E}(t_k^{(i)}),
\end{equation}
where $M_i$ denotes the number of descriptor sentences produced for 
$\mathcal{D}_i$ by Algorithm~\ref{alg:eda_embedding}. The collection 
$\Phi_i$ is represented as an embedding matrix $V_i \in 
\mathbb{R}^{ d_{\mathcal{E}} \times M_i }$ formed by concatenating the columns, 
$[\mathbf{v}_1^{(i)}; \cdots; \mathbf{v}_{M_i}^{(i)}]$
 which serves as the basis for 
cross-dataset comparison.

Cross-dataset similarity is quantified by comparing embedding matrices 
using Canonical Correlation Analysis (CCA). Given two embedding 
matrices $V_i \in \mathbb{R}^{ d_{\mathcal{E}} \times M_i }$ and $V_j 
\in \mathbb{R}^{d_{\mathcal{E}} \times M_j}$, CCA identifies $r^*$ pairs 
of canonical directions that maximize the correlation between the 
projected embeddings, yielding canonical correlations $\rho_1 \geq 
\rho_2 \geq \cdots \geq \rho_r \geq 0$, where $r^* = \min(\hat{r}_i, \hat{r}_j)$. The scalar similarity between $\mathcal{D}_i$ 
and $\mathcal{D}_j$ is defined as the mean canonical correlation:
\begin{equation}
    s_{ij} = \frac{1}{r} \sum_{l=1}^{r} \rho_l,
\end{equation}
and the corresponding distance is $D_{ij} = 1 - s_{ij}$, with 
$D_{ii} = 0$. This pairwise computation is repeated across all 
$\binom{N}{2}$ dataset pairs in the catalog to construct the distance 
matrix $D \in \mathbb{R}^{N \times N}$, which is then used as the 
basis for hierarchical clustering. Ward D2 linkage is adopted, 
which can be modified based on the specific application, as the 
agglomeration criterion, producing a dendrogram $\mathcal{T}$ that 
organizes the catalog by statistical similarity and supports efficient 
retrieval when a new dataset $\mathcal{D}_q$ is introduced.

To recover interpretable alignment between dataset embedding spaces, a 
penalized formulation of CCA is applied using the penalized matrix 
decomposition framework~\citep{witten2009penalized}. An $\ell_1$ norm 
penalty is imposed on the canonical weight vectors $w_i \in 
\mathbb{R}^{d_{\mathcal{E}}}$ and $w_j \in \mathbb{R}^{d_{\mathcal{E}}}$, 
inducing sparsity in the solution. Specifically, the penalized CCA 
problem is:
\begin{equation}
    \max_{w_i,\, w_j} \; w_i^\top V_i^\top V_j w_j
    \quad \text{subject to} \quad
    \|w_i\|_2 \leq 1, \;
    \|w_j\|_2 \leq 1, \;
    \|w_i\|_1 \leq \lambda, \;
    \|w_j\|_1 \leq \lambda,
\end{equation}
where $\lambda > 0$ is a sparsity penalty selected either via sample 
permutation or fixed by the practitioner. Because each dimension of 
$V_i$ corresponds to a distinct descriptor sentence $t_k^{(i)}$ rather 
than an arbitrary linear combination of features, the nonzero entries 
of the sparse solution $w_i^* \in \mathbb{R}^{d_{\mathcal{E}}}$ and 
$w_j^* \in \mathbb{R}^{d_{\mathcal{E}}}$ identify which specific 
descriptors or variable-level quantities drive cross-dataset 
correlation. This provides a degree of interpretability that is not 
available under standard CCA, where the rank-$r$ approximation 
distributes weight across all embedding dimensions simultaneously. The 
penalized CCA problem is solved iteratively for components $1, \ldots, 
r$, deflating $V_i$ and $V_j$ after each component by
subtracting the rank-one contribution of each solved component prior to estimating the next.

\begin{figure}[H]
    \centering
    \includegraphics[width=0.99\linewidth]{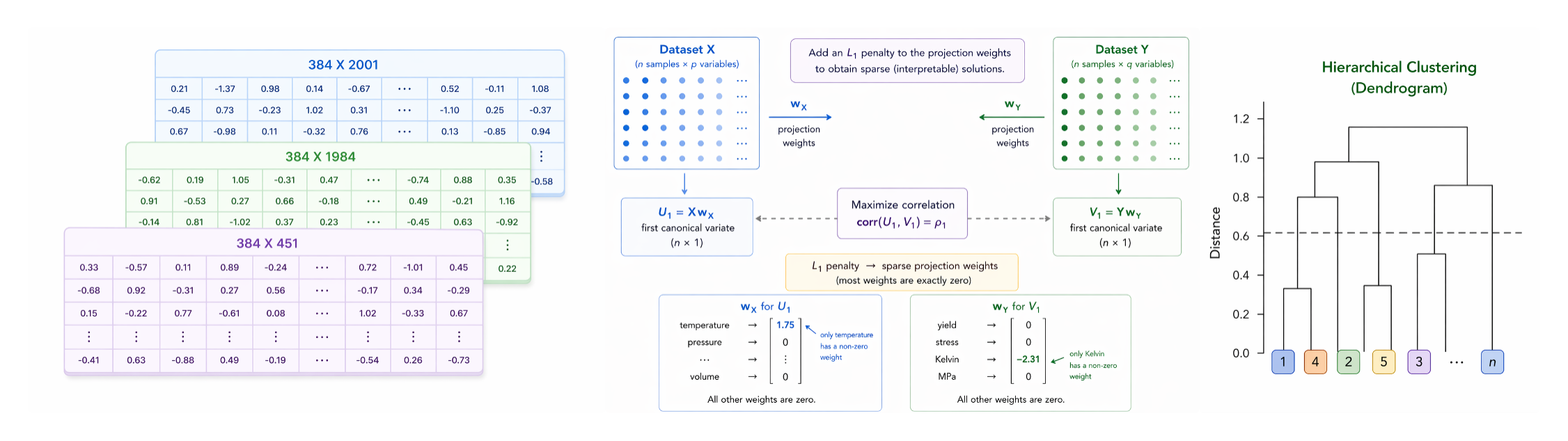}
    \caption{Interpretable embedding process flow for determining data 
             similarities.}
    \label{fig:embedding_similarity}
\end{figure}
\begin{algorithm}[H]
\caption{Embedding, CCA Similarity, Retrieval, and Interpretable 
         Alignment}
\label{alg:cca_retrieval}
\begin{algorithmic}[1]

\Require Sentence collections $\{\mathcal{S}_1, \ldots, \mathcal{S}_N\}$ 
         from Algorithm~\ref{alg:eda_embedding} for catalog datasets 
         $\{\mathcal{D}_1, \ldots, \mathcal{D}_N\}$;
         query collection $\mathcal{S}_q$ for new dataset 
         $\mathcal{D}_q$;
         sentence transformer $\mathcal{E}: \mathcal{T} \rightarrow 
         \mathbb{R}^{d_{\mathcal{E}}}$;
         number of neighbors $k \in \mathbb{Z}^+$;
         linkage $\ell$;
         interpretability flag 
         $\texttt{pcca} \in \{\texttt{true}, \texttt{false}\}$;
         penalty $\lambda > 0$ (if $\texttt{pcca} = \texttt{true}$)

\Ensure Top-$k$ neighbors of $\mathcal{D}_q$; distance matrix 
        $D \in \mathbb{R}^{N \times N}$; dendrogram $\mathcal{T}$; 
        sparse weights $\{(w_i^*, w_j^*)\}$ 
        (if $\texttt{pcca} = \texttt{true}$)

\vspace{0.5em}
\State \textit{// Embedding}

\For{each $\mathcal{S}_i$, $i = 1,\ldots,N$ and query 
     $\mathcal{S}_q$}
    \State Encode each $t_k \in \mathcal{S}_i$ via $\mathcal{E}$; 
           stack rows to form 
           $V_i \in \mathbb{R}^{ d_{\mathcal{E}} \times M_i }$
\EndFor

\vspace{0.5em}
\State \textit{// Pairwise CCA Similarity and Distance Matrix}

\For{each pair $(i,j)$ with $1 \leq i < j \leq N$}
    \State Mean-center and $\ell_2$-normalize $V_i$, $V_j$; 
           apply Tikhonov regularization $\alpha = 10^{-6}$
    \State Compute CCA; set 
           $s_{ij} = \frac{1}{r}\sum_{l=1}^{r}\rho_l$ where 
           $r = \min(M_i, M_j, d_{\mathcal{E}}) - 1$;
           set $D_{ij} = D_{ji} = 1 - s_{ij}$
\EndFor
\State Set $D_{ii} = 0$ for all $i$

\vspace{0.5em}
\State \textit{// Hierarchical Clustering and Retrieval}

\State Apply hierarchical clustering to $D$ with linkage $\ell$; 
       produce $\mathcal{T}$

\For{each $\mathcal{D}_i$ in catalog}
    \State Compute $d_{qi} = 1 - \frac{1}{r}
           \sum_{l=1}^{r}\rho_l$ between $V_q$ and $V_i$
\EndFor
\State Sort by $d_{qi}$ ascending; return top $k$ as nearest 
       neighbors of $\mathcal{D}_q$

\vspace{0.5em}
\State \textit{// Penalized CCA (conditional)}

\If{$\texttt{pcca} = \texttt{true}$}
    \For{each dataset pair $(\mathcal{D}_i, \mathcal{D}_j)$}
        \State Estimate $\lambda$ via permutation if not 
               specified~\citep{witten2009penalized}
        \State Solve $\max_{w_i, w_j} w_i^\top V_i^\top V_j w_j$ 
               s.t. $\|w_i\|_2, \|w_j\|_2 \leq 1$, 
               $\|w_i\|_1, \|w_j\|_1 \leq \lambda$; 
               obtain $w_i^*, w_j^* \in \mathbb{R}^{d_{\mathcal{E}}}$
        \State Map nonzero indices of $w_i^*, w_j^*$ to 
               descriptor sentences and variable names
        \State Deflate $V_i$, $V_j$; repeat for 
               components $2,\ldots,r$
    \EndFor
\EndIf

\State \Return top-$k$ neighbors; $D$; $\mathcal{T}$; 
       $\{(w_i^*, w_j^*)\}$

\end{algorithmic}
\end{algorithm}

\section{Experimental}\label{sec:datasources}

\subsection{Experimental Setup}
\label{sec:setup}

All experiments were conducted on an Apple MacBook Pro with an Apple M3 
chip and 36 GB unified memory, running macOS. Statistical descriptor 
computation was performed in R using the \texttt{mungeR} 
package~\citep{kunz2026mungeR}. Sentence embeddings were generated using the 
\verb|all-MiniLM-L6-v2| sentence transformer~\citep{wang2020minilm}, 
accessed via the \texttt{sentence-transformer} package in Python \citep{reimers-2019-sentence-bert}, 
producing 384-dimensional embedding vectors for each descriptor sentence.

CCA was computed using the standard singular value decomposition 
formulation. To ensure numerical stability, embedding matrices were 
mean-centered and $\ell_2$-normalized prior to CCA computation. The number of 
canonical components $r$ was set to
$\min(R_i, R_j)$ for each dataset pair, where $R_i$ and $R_j$ 
denote the estimated numerical rank, via Gavish--Donoho, for datasets $\mathcal{D}_i$ 
and $\mathcal{D}_j$ respectively.

Penalized CCA was implemented using the \texttt{PMA} package in 
R~\citep{witten2009penalized}. Penalty parameters were selected via 
sample permutation with 50 permutation replicates for the 
permutation-tuned condition, and fixed at $\lambda = 10^{-6}$ for the 
illustrative fixed-penalty condition. Initialization instability in the 
PMA coordinate descent algorithm was mitigated by running five random 
initializations per component and retaining the solution with the highest 
canonical correlation. Hierarchical clustering was performed using Ward 
D2 linkage via the \texttt{hclust} function in R. All random seeds were 
set to 42 throughout.

\subsection{Datasets}\label{sec:datasets}

Three categories of datasets were selected for evaluation based on open-source availability and domain relevance: general-purpose benchmark datasets from the UCI Machine Learning Repository, steel and materials property datasets sourced from Citrine Informatics and the Materials Project, and nuclear-grade graphite characterization data maintained within the Nuclear Data Management and Analysis System (NDMAS) at Idaho National Laboratory. Summary statistics for each dataset are provided in the accompanying tables.

Three datasets from the UCI Machine Learning Repository \citep{anderson1936species,fisher1936use,iris_53, heck1998corsika,magic_gamma_telescope_159, cortez2009modeling,wine_quality_186}
 were included to establish baseline performance on well-characterized benchmark datasets. The Iris dataset contains 150 specimens described by four morphological measurements (sepal length, sepal width, petal length, and petal width) across three species of \textit{Iris}, making it a canonical low-dimensional classification benchmark. The Wine Quality datasets consist of separate red and white wine tables, each recording physicochemical properties such as acidity, residual sugar, chlorides, and sulfur dioxide levels alongside a sensory quality score; the red and white tables are treated as independent datasets in this work to preserve differences in their marginal distributions and sample sizes. The MAGIC Gamma Telescope dataset contains simulated high-energy gamma particle shower measurements recorded by an imaging atmospheric Cherenkov telescope, with features derived from the spatial and intensity distribution of the Cherenkov light pattern; the classification target distinguishes genuine gamma signal events from hadronic background events.

Four datasets characterizing the mechanical and physical properties of steels and inorganic materials were included to evaluate performance in a materials informatics context \citep{agrawal2014exploration,citrination_agrawal, dunn2020benchmarking, citrination_mechanical_properties, ward2016general, de1988cohesion}. These datasets vary in size, feature construction strategy, and property target, providing complementary coverage of the materials property prediction problem. The Matbench steels and Citrine steels datasets originate from the same underlying experimental records; the Matbench version applies a standardized data cleaning and cross-validation protocol consistent with the broader Matbench benchmark suite, while the Citrine version represents the data in its original deposited form. Both are treated as separate entries to preserve this distinction.

Eight datasets characterizing nuclear-grade graphite were extracted from NDMAS at Idaho National Laboratory, as documented in \cite{pham2025summary,NDMAS_Graphite}. The data support qualification of graphite grades for use as structural core components in high-temperature gas-cooled reactors (HTGRs). Specimens are drawn primarily from the 2114, IG-110, NBG-17, NBG-18, and PCEA grades, tested under both baseline (unirradiated) and irradiated conditions through the Advanced Graphite Capsule (AGC) irradiation program. Each of the eight NDMAS graphite datasets is treated as an independent tabular dataset, analyzed separately to reflect the practical scenario in which individual property tables are ingested and characterized without cross-table joins or shared identifiers.

A full list of the variables within each dataset is described in Section~\ref{sec:supplementary}.

\section{Results}\label{sec:results}
The experimental validation is organized into three subsections. 
Section~\ref{sec:results:set1} evaluates nearest-neighbor 
retrieval accuracy across embedding conditions and baseline methods, 
establishing whether the proposed EDA-derived embeddings recover 
meaningful dataset similarity structure at the individual dataset level. 
Section~\ref{sec:results:set2} expands this analysis to the full 
pairwise distance structure through hierarchical clustering, examining 
how embedding condition and differential privacy budget jointly affect 
the organization of the complete dataset collection. 
Section~\ref{sec:results:set3} demonstrates the interpretability of 
the framework through penalized CCA, identifying which statistical 
descriptors and variable-level quantities drive cross-dataset alignment 
for selected dataset pairs. Across all three analyses, results are 
reported for the full univariate and multivariate embedding condition, 
the multivariate-only ablation, and differentially private embeddings 
at $\epsilon \in \{0.1, 1, 10\}$.

\subsection{First neighbor relations and accuracy}\label{sec:results:set1}

A nearest-neighbor scheme analogous to document similarity is used to assess similarity between tabular datasets. Each dataset is compared to its nearest neighbor under several embedding conditions: the Uni/Multi-variate embedding (as described in the methodology); the Multivariate embedding, which summarizes univariate correlations, dimensionality, singular values, and multivariate responses; and differentially private embeddings at varying noise levels, where $\epsilon = 0.1$ corresponds to the highest noise injection and $\epsilon = 10$ to the lowest. Results are organized by dataset group: Table~\ref{tab:correlation_general} for the UCI Machine Learning datasets, Table~\ref{tab:correlation_materials} for the materials informatics datasets, and Tables~\ref{tab:correlation_ndmas1} and~\ref{tab:correlation_ndmas2} for the NDMAS graphite datasets. This section focuses on the nearest neighbor assignment; broader interpretation of embedding similarity is explored in Section~\ref{sec:results:set3}.

Among the UCI datasets (Table~\ref{tab:correlation_general}), the red and white wine tables serve as a natural control pair, as both describe the same physicochemical feature space despite differing sample sizes. 
Considering them as a pair, the $P@1$ score, i.e., the correct top-1 nearest neighbors over all queries, results in a score of 0.90.
The Iris and MAGIC Gamma Telescope datasets are included to characterize variation in nearest-neighbor assignment across less related datasets. Under the Multivariate embedding, the Iris and red wine datasets are assigned unexpected nearest neighbors relative to the other embedding conditions, suggesting that matrix-level summaries alone may not carry sufficient information to recover meaningful similarity. Across all other embedding conditions, Iris is consistently matched to IRRSummary, which is notable because if the similarity were driven purely by shared dimensionality, the Multivariate embedding would be expected to produce the same assignment. The telescope dataset exhibits the greatest instability in nearest-neighbor assignment across embedding spaces, reflecting its dissimilarity from the remaining datasets in this group.

Among the materials informatics datasets (Table~\ref{tab:correlation_materials}), the two Citrine datasets serve as a control pair, as both represent alloy compositions encoded as weight percentages. 
This pairing was recovered as mutual nearest neighbors across all embedding conditions, including at $\epsilon = 0.1$, demonstrating robustness to substantial noise injection and a perfect $P@1$ score. 
The Matbench steels dataset, which uses different units and measurement conventions than the Citrine steels dataset, is nonetheless correctly matched to Citrine steels in all conditions except the Multivariate embedding. 
Similarly, the Magpie and Miedema datasets span different feature spaces but share a common materials property description context, and are matched as nearest neighbors in all conditions except Multivariate resulting in a $P@1$ score of 0.8. 
The Miedema dataset shows the most variation within this group, with Tensile as its most frequent nearest neighbor outside the Magpie match; this may reflect the overlap in compressibility, shear modulus, melting point, and structural stability features shared between the two datasets.

\begin{sidewaystable}[htbp]
    \centering
    \caption{First nearest neighbors for the UCI Machine Learning datasets.}\label{tab:correlation_general}
    \begin{tabular*}{\textheight}{c | cccc}
    Dataset & iris & telescope & wine red & wine white\\
    \hline
    Univariate and Multivariate         & IRRSummary &     Flex & wine white & wine red\\
    Multivariate      & \textbf{Flex} &     iris & \textbf{matbench steels} & wine red\\
    $\epsilon = 0.1$ & IRRSummary &     iris & wine white & wine red\\
    $\epsilon = 1$   & IRRSummary & SpecNeut & wine white & wine red\\
    $\epsilon = 10$  & IRRSummary &     Flex & wine white & wine red\\
    \hline
    \end{tabular*}

    \vspace{4em}

    \caption{First nearest neighbors for the materials informatics datasets.}\label{tab:correlation_materials}
    \begin{tabular*}{\textheight}{c | ccccc}
    Dataset                             & citrine agrawal & citrine steel & magpie & matbench steels & miedema\\
    \hline
    Univariate and Multivariate         & citrine steel & citrine agrawal &          miedema & citrine steel &  Tensile\\
    Multivariate                        & citrine steel & citrine agrawal & \textbf{matbench steels} &   \textbf{wine red} & telescope\\
    $\epsilon = 0.1$                    & citrine steel & citrine agrawal &          miedema & citrine steel &     iris\\
    $\epsilon = 1$                      & citrine steel & citrine agrawal &          miedema & citrine steel &  wine red\\
    $\epsilon = 10$                     & citrine steel & citrine agrawal &          miedema & citrine steel &  Tensile\\
    \hline
    \end{tabular*}

     \vspace{4em}

    \caption{First nearest neighbors for the NDMAS datasets without ambient temperature and humidity variables.}
    \label{tab:correlation_ndmas1}
    \begin{tabular*}{\textheight}{c | ccccc}
    Dataset          & GasChem & IRRSummary & SpecNeut & SpecThermal & ThermalConductivity\\
    \hline
    Univariate and Multivariate        &   IRRSummary &        SpecNeut & IRRSummary & ThermalConductivity & SpecThermal\\
    Multivariate      &   IRRSummary & \textbf{iris}   & IRRSummary & ThermalConductivity & SpecThermal\\
    $\epsilon = 0.1$ & \textbf{iris} & \textbf{iris}  & IRRSummary & ThermalConductivity & SpecThermal\\
    $\epsilon = 1$   &   IRRSummary &        SpecNeut & IRRSummary & ThermalConductivity & SpecThermal\\
    $\epsilon = 10$  &    Resonance &        SpecNeut & IRRSummary & ThermalConductivity & SpecThermal\\
    \hline
    \end{tabular*}

    \vspace{4em}

    \caption{First nearest neighbors for the NDMAS datasets with ambient temperature and humidity variables.}
    \label{tab:correlation_ndmas2}
    \centering
    \begin{tabular*}{\textheight}{c| ccccccc} 
    Dataset          & BulkDensity            & Compression & Flex        & Resistivity               & Resonance       & SonicVelocity             & Tensile\\
    \hline
    Univariate and Multivariate         & Resonance              & Flex        & Compression & SonicVelocity             & SonicVelocity   & Resistivity               & Compression\\
    Multivariate      & \textbf{SonicVelocity} & Flex        & Compression & SonicVelocity             & SonicVelocity   & Resistivity               & Compression\\
    $\epsilon = 0.1$ & Resonance              & Tensile     & Compression & \textbf{Resonance}        & SonicVelocity   & \textbf{Resonance}        & Compression\\
    $\epsilon = 1$   & Resonance              & Tensile     & Compression & SonicVelocity             & SonicVelocity   & \textbf{Resonance}        & Compression\\
    $\epsilon = 10$  & Resonance              & Tensile     & Compression & SonicVelocity             & SonicVelocity   & Resistivity               & Compression\\
    \hline
    \end{tabular*}
\end{sidewaystable}

The NDMAS nearest-neighbor results are presented in two tables (Tables~\ref{tab:correlation_ndmas1}, \ref{tab:correlation_ndmas2}) separated by whether datasets include ambient temperature and humidity variables. Within the datasets lacking these variables (Table~\ref{tab:correlation_ndmas1}), GasChem, IRRSummary, and SpecThermal are the most mutually similar, likely reflecting shared operational feature spaces including temperature, concentration, fluence, and displacements per atom (dpa). Under high obfuscation or Multivariate conditions, IRRSummary and GasChem are displaced toward Iris as a nearest neighbor, indicating that these embeddings no longer retain enough structure to resolve within-group similarity. Within the datasets containing ambient temperature and humidity (Table~\ref{tab:correlation_ndmas2}), BulkDensity, Resistivity, Resonance, and SonicVelocity show overlapping nearest-neighbor assignments, particularly under obfuscation and Multivariate conditions. This cross-assignment pattern suggests that the full similarity structure within this group warrants examination of the complete similarity matrix, including hierarchical clustering, rather than nearest-neighbor summaries alone.

When considering all dataset pairs, i.e., red and white wines, citrine datasets, magpie connecting to miedema, and matbench steel to citrine steels, the total $P@1$ retrieval accuracy score was 0.9.

\subsection{Hclust results}\label{sec:results:set2}

The previous section established that using only multivariate information or a high degree of obfuscation can introduce variation in nearest-neighbor assignment. This section expands on that analysis by examining the full pairwise distance structure, $D_{ij}$, across all datasets. Hierarchical clustering over these distances provides a means to identify which datasets should be grouped together and, more importantly, how cluster-level embeddings can support retrieval: a new dataset can be compared to the catalog by computing CCA between its embedding space and a cluster embedding space, rather than performing dataset-to-dataset comparisons.

\begin{figure}[htbp]
  \centering
\begin{subfigure}{.45\linewidth}
  \includegraphics[width=\linewidth]{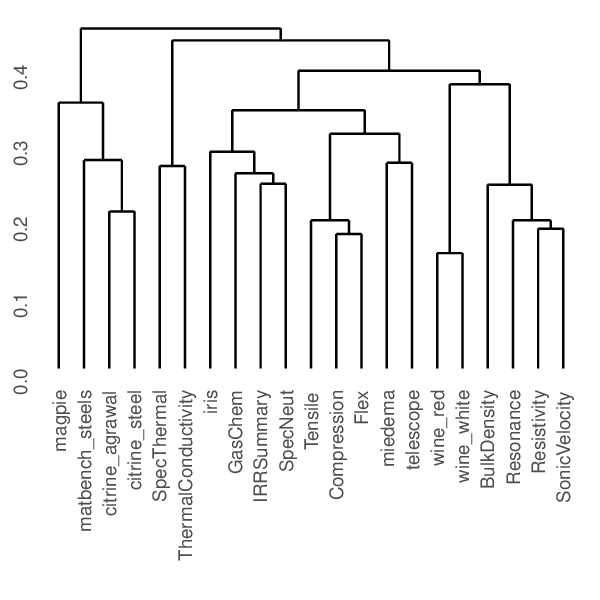}
  \caption{Univariate and multivariate}
  \label{fig:hclustAvailableDataA}
\end{subfigure}
~
\begin{subfigure}{.45\linewidth}
  \includegraphics[width=\linewidth]{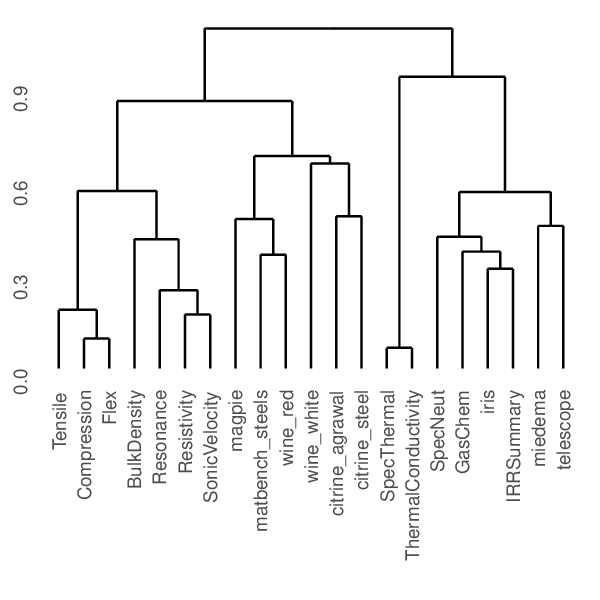}
  \caption{Multivariate only}
  \label{fig:hclustAvailableDataD}
\end{subfigure}

\medskip
\begin{subfigure}{.45\linewidth}
  \includegraphics[width=\linewidth]{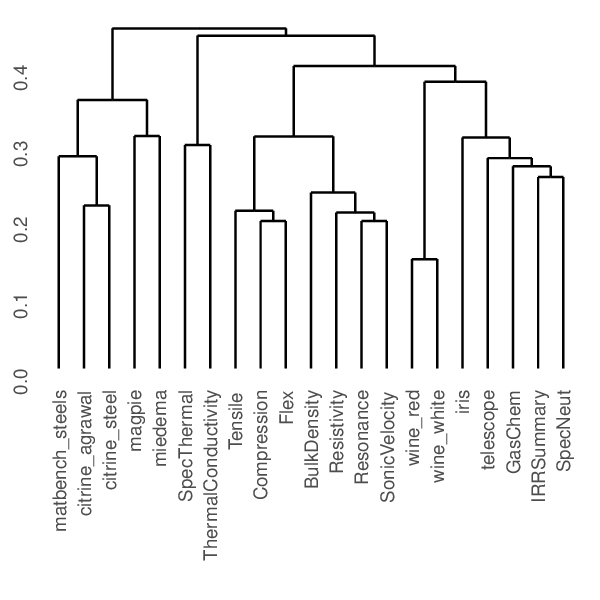}
  \caption{Univariate and multivariate, excluding statistic value}
  \label{fig:hclustAvailableDataB}
\end{subfigure}
\begin{subfigure}{.45\linewidth}
  \includegraphics[width=\linewidth]{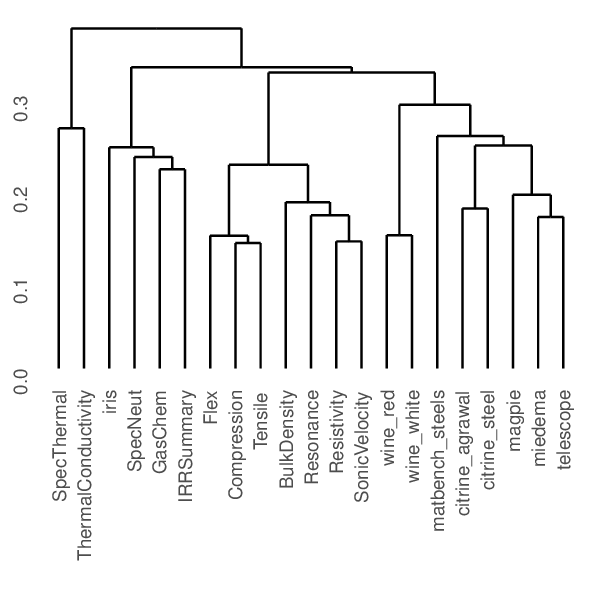}
  \caption{Univariate and multivariate, excluding statistic value and variable name}
  \label{fig:hclustAvailableDataC}
\end{subfigure}
\caption{Hierarchical clustering (Ward D2) across four embedding conditions.}
\label{fig:hclustAvailableData}
\end{figure}

Figure~\ref{fig:hclustAvailableData} compares hierarchical clustering results (Ward D2 linkage) across four embedding conditions. Subfigure~\ref{fig:hclustAvailableDataA} uses the full univariate and multivariate descriptor set. Subfigure~\ref{fig:hclustAvailableDataD} uses only multivariate descriptors, isolating the effect of excluding univariate content. Subfigures~\ref{fig:hclustAvailableDataB} and~\ref{fig:hclustAvailableDataC} progressively ablate contextual information from the sentence descriptors: first removing the statistic value (e.g., reducing ``The median is 5.8. Variable: sepal.length'' to the variable name alone), then also removing the variable name.

In the full univariate and multivariate dendrogram (Figure~\ref{fig:hclustAvailableDataA}), the only unexpected grouping relative to the nearest-neighbor results is the pairing of the Miedema and telescope datasets, which is consistent with the high nearest-neighbor variability observed for both in Section~\ref{sec:results:set1}. When restricted to multivariate descriptors only (Figure~\ref{fig:hclustAvailableDataD}), the cluster structure is largely preserved, with the primary change being that the red and white wine datasets are no longer directly linked, though they remain within the same cluster. Excluding the statistic value from the sentence descriptor (Figure~\ref{fig:hclustAvailableDataB}) yields tighter grouping of the Magpie, Miedema, and telescope datasets relative to the full descriptor condition. Further excluding the variable name (Figure~\ref{fig:hclustAvailableDataC}) produces a cluster structure visually similar to the full univariate and multivariate result, but with reduced inter-cluster separation as reflected in the dendrogram heights. Together, these results indicate that variable names and statistic values improve cluster resolution but are not required to recover the broad cluster structure, a property that is particularly relevant for datasets with nonstandard schemas or inconsistent naming conventions.

\begin{figure}[htbp]
  \centering
\begin{subfigure}{.45\linewidth}
  \includegraphics[width=\linewidth]{plots/munger_embeddings.eps}
  \caption{No differential privacy}
  \label{fig:hclustDPA}
\end{subfigure}
~
\begin{subfigure}{.45\linewidth}
  \includegraphics[width=\linewidth]{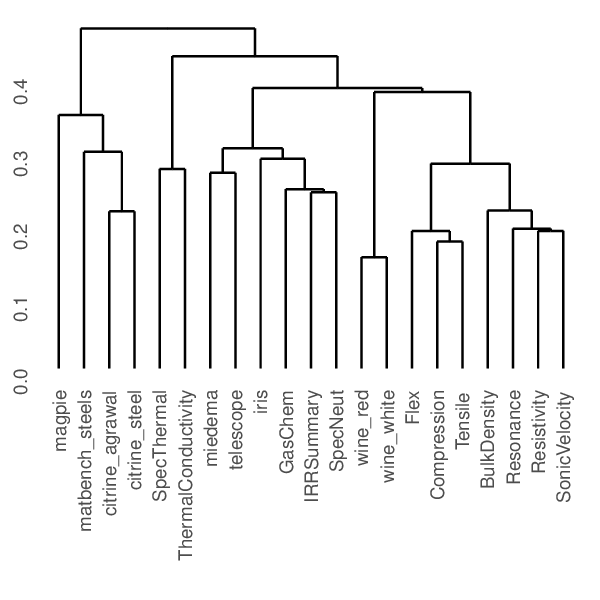}
  \caption{$\epsilon = 10$}
  \label{fig:hclustDPB}
\end{subfigure}

\medskip
\begin{subfigure}{.45\linewidth}
  \includegraphics[width=\linewidth]{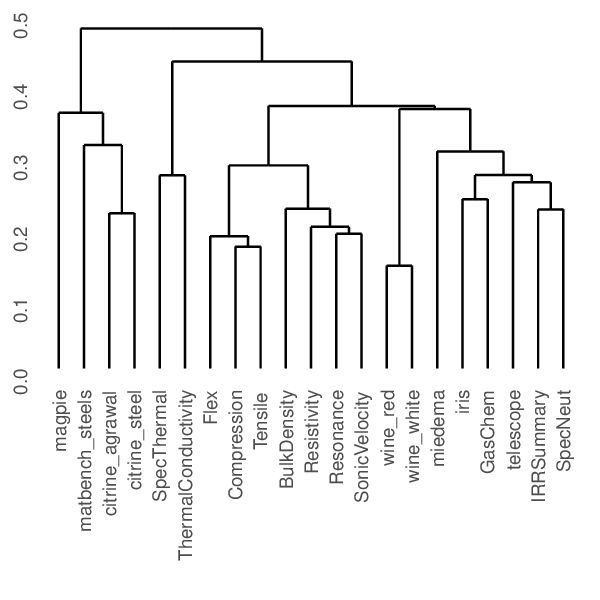}
  \caption{$\epsilon = 1$}
  \label{fig:hclustDPC}
\end{subfigure}
\begin{subfigure}{.45\linewidth}
  \includegraphics[width=\linewidth]{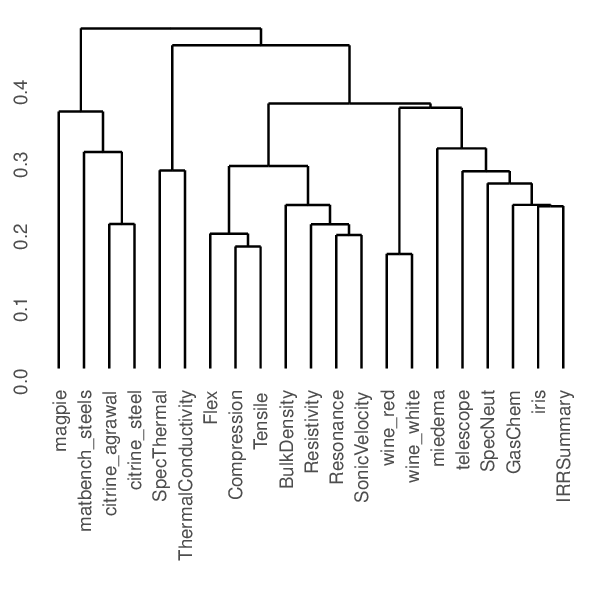}
  \caption{$\epsilon = 0.1$}
  \label{fig:hclustDPD}
\end{subfigure}
\caption{Hierarchical clustering (Ward D2) under varying differential privacy budgets.}
\label{fig:hclustDP}
\end{figure}

The effect of differential privacy on cluster structure is examined in 
Figure~\ref{fig:hclustDP}, which compares dendrograms across four privacy 
budgets using the full univariate and multivariate descriptor set. Recall 
that $\epsilon$ controls the privacy budget, with smaller values providing 
stronger privacy guarantees at the cost of greater noise injection. 
Comparing the no-privacy baseline (Figure~\ref{fig:hclustDPA}) to the 
lowest noise condition ($\epsilon = 10$, Figure~\ref{fig:hclustDPB}), 
the primary structural change is the migration of the Tensile, 
Compression, and Flex datasets toward closer association with Resonance 
and Resistivity, rather than with the operational parameter datasets 
GasChem and IRRSummary. To understand why specific dataset pairs are more 
susceptible to this structural displacement than others, Table~\ref{tab:spectral_entropy} 
examines the spectral entropy of the singular value spectrum for a subset 
of NDMAS datasets across increasing levels of differential privacy noise, 
where spectral entropy serves as a measure of how uniformly variance is 
distributed across the directions of the data matrix.
As the privacy budget decreases, the spectral entropy of Resonance and Resistivity converges toward that of Tensile, Compression, and Flex, while IRRSummary, GasChem, and SpecNeut remain near one. This convergence accounts for the dendrogram shifts observed for BulkDensity, Resonance, Resistivity, and SonicVelocity under obfuscation. More broadly, these results indicate that differential privacy budget selection will affect the similarity structure of datasets whose embeddings are dominated by variance descriptors; however, the overall cluster organization remains relatively stable, reflecting the inherent measurement noise present in the underlying data.

\begin{sidewaystable}[htbp]
  \centering
  \caption{Spectral entropy for a subset of NDMAS datasets under increasing differential privacy noise. As the privacy budget decreases, spectral entropy converges across datasets, shifting the similarity structure toward variance-dominated groupings.}
  \label{tab:spectral_entropy}
  \begin{tabular*}{\textheight}{l | ccc | ccc | ccc}
    & BulkDensity & Resonance & SonicVelocity & Compression & Flex & Tensile & GasChem & IRRSummary & SpecNeut\\
    \hline
    No DP         & 1.401 & 0.989 & 1.592 & 1.609 & 1.875 & 1.951 & 0.998 & 0.999 & 0.155\\
    $\epsilon = 10$ & 1.562 & 0.994 & 1.974 & 1.979 & 1.994 & 2.802 & 0.998 & 1.000 & 0.838\\
    $\epsilon = 1$  & 1.585 & 0.999 & 1.999 & 1.999 & 1.999 & 2.807 & 0.989 & 1.000 & 0.999\\
    $\epsilon = 0.1$ & 1.585 & 1.000 & 2.000 & 1.999 & 1.999 & 2.806 & 1.000 & 1.000 & 1.000\\
  \end{tabular*}

\end{sidewaystable}

\subsection{CCA results}\label{sec:results:set3}

To ensure interpretability of dataset associations, an $\ell_1$ penalty is applied to the CCA calculation via penalized matrix decomposition. This yields sparse canonical loadings that identify the key variables contributing to each component. Three components are used here for demonstration; in practice, the number of components should be selected by examining the full cross-correlation structure.

Table~\ref{tab:cca_permutation} presents penalized CCA results between the two Citrine datasets, where the penalty parameter was selected via sample permutation~\citep{witten2009penalized}. Despite the two datasets using different units and column naming conventions (atomic percent vs. weight percent) to describe alloy composition, the first component recovers a shared set of alloying elements across the two representations. The specific elements selected do not correspond exactly by name, which warrants follow-up with a subject matter expert familiar with both datasets. The second component loads on multivariate matrix statistics in both datasets, which may reflect shared variance structure among the composition variables. The third component aligns on measures of mechanical strength, with fatigue strength in the Citrine Agrawal dataset corresponding to yield and ultimate tensile strength in the Citrine steels dataset. This is an encouraging result given that both studies examined the properties of steel compositions, and confirms that the penalized CCA recovers physically meaningful cross-dataset structure.

To illustrate the interpretability of penalized CCA more broadly, Table~\ref{tab:penalizedCCA} presents results for three dataset pairs under a fixed penalty of $\lambda = 10^{-6}$: Citrine steel vs. Citrine Agrawal, Magpie vs. Miedema, and Tensile vs. Miedema. This fixed penalty is used for demonstration only; in practice the penalty should be optimized per dataset pair.

For the Citrine pair, restricting the penalty concentrates the first component on manganese content and the third component specifically on ultimate tensile and fatigue strength. For the Magpie and Miedema comparison, the first component links the number of elements to enthalpy of transformation, which have no direct physical relationship; however, the second component recovers a known correspondence between atomic weight and electron density, and the third component aligns molar volume across both datasets. The Tensile and Miedema comparison yields the most physically interpretable result: the first two components both align on elastic and shear modulus, and the third component relates pressure at fracture to melting point. These examples confirm that penalized CCA can surface interpretable cross-dataset variable correspondences even under a fixed, non-optimized penalty parameter.

\begin{sidewaystable}[htbp]
    \centering

    \caption{Penalized CCA results between the Citrine datasets. The $\ell_1$ penalty parameter was selected via permutation~\citep{witten2009penalized}.}\label{tab:cca_permutation}
    \begin{tabular*}{\textheight}{c : c}
    citrine agrawal & citrine steel\\
    \hline
    C at\%, Cr at\%, Fe at\%, Mn at\%, Ni at\%, Si at\% & Co wt\%, Cr wt\%, Mn wt\%, Mo wt\%, Ni wt\%, Si wt\%, Ti wt\%, V wt\%\\
    \hdashline
    matrix & matrix\\
    \hdashline
    Fatigue Strength (MPa) & Yield Strength (MPa), Ultimate Tensile Strength (MPa)
    \end{tabular*}

    \vspace{4em}

    \caption{Non-zero canonical loadings from penalized CCA for three dataset pairs under a fixed penalty of $\lambda = 10^{-6}$. This penalty is illustrative; optimal selection should be performed per dataset pair.}
    \label{tab:penalizedCCA}
    \begin{tabular*}{\textheight}{c | c : c | c : c | c : c}
    Dataset & citrine steel & citrine agrawal & magpie & miedema & Tensile & miedema\\
    \hline
    Component 1 & Mn wt\% & Mn at\% & Number & H trans & elastic modulus tensile & shear modulus\\
    \hdashline
    Component 2 & matrix & matrix & Atomic Weight & electron density & elastic modulus tensile & shear modulus\\
    \hdashline
    Component 3 & UTS (MPa) & Fatigue Strength (MPa) & Molar Volume & molar volume & $p$ at break & melting point    
\end{tabular*}

\end{sidewaystable}

\section{Conclusion}\label{sec:conclusion}

\subsection{Summary and Key Findings}

This work introduced a framework for representing numeric tabular datasets 
as statistically grounded embeddings suitable for retrieval-augmented 
generation and cross-dataset comparison. The approach characterizes each 
dataset through a structured EDA pipeline that computes univariate and 
multivariate descriptors, serializes those descriptors into natural 
language sentences, and encodes them into a shared vector space via a 
pretrained sentence transformer. Pairwise dataset similarity is then 
quantified through CCA, with a penalized formulation providing sparse, 
interpretable canonical loadings that identify which specific variables 
or matrix-level quantities drive cross-dataset alignment.

Evaluation across 15 datasets spanning general-purpose benchmarks, 
materials informatics, and nuclear-grade graphite characterization 
demonstrated that the proposed embeddings recover meaningful similarity 
structure across substantially different domains and feature spaces. 
Nearest-neighbor retrieval achieved $P@1$ of 0.9 across all embedding conditions. Cluster 
structure was largely preserved under differential privacy noise, with 
perturbation effects concentrated among datasets whose embeddings are 
dominated by variance descriptors, as quantified through spectral entropy. 
Penalized CCA recovered physically interpretable cross-dataset 
correspondences without requiring shared variable names or feature 
conventions, including alignment of alloying elements across atomic and 
weight percent representations and correspondence of mechanical strength 
measures across independently constructed steel datasets. Together, these 
results demonstrate that EDA-derived statistical fingerprints provide a 
compact, semantically interpretable basis for organizing heterogeneous 
numeric data within LLM-compatible embedding spaces.

\subsection{Limitations and Future Work}

Several limitations of the current framework warrant consideration. 
Embedding quality is bounded by the statistical content of the EDA 
descriptors, which summarize marginal and joint distributions but do not 
capture higher-order interactions or domain-specific structure falling 
outside the descriptor vocabulary. The sentence transformer used for 
encoding, all-MiniLM-L6-v2, was pretrained on general text corpora and 
has not been fine-tuned for statistical or scientific language; 
domain-adapted encoders may improve retrieval precision in specialized 
settings. The penalized CCA results presented here use fixed or 
permutation-selected penalty parameters applied uniformly across 
components, and a more principled cross-validated tuning procedure would 
strengthen the interpretability guarantees. Finally, the evaluation relies 
on nearest-neighbor and clustering metrics as proxies for retrieval 
quality; direct evaluation in an end-to-end RAG pipeline with downstream 
task performance would provide stronger empirical grounding.

A particularly promising direction for future work is the application of 
the proposed similarity measures to algorithm and model selection for 
unknown datasets, connecting the framework to the AutoML and 
meta-learning literature. In meta-learning, algorithm performance on a 
new task is estimated from the performance of candidate algorithms on 
similar prior tasks, with similarity typically defined over hand-crafted 
dataset meta-features such as dimensionality, class balance, and 
statistical moments \cite{vanschoren2018meta, feurer2015efficient}. The EDA-derived embedding proposed here 
provides a richer and more principled characterization of dataset 
structure than existing meta-feature sets, and the CCA-based similarity 
measure offers a direct mechanism for identifying which prior datasets 
are most informative for warm-starting algorithm selection. In the AutoML 
context, this enables a retrieval-based initialization strategy: given an 
unfamiliar dataset, the nearest neighbors in embedding space identify 
candidate algorithms or hyperparameter configurations whose assumptions 
are most compatible with the observed statistical structure of the new 
data, reducing the search space before any model is trained. For example, 
a dataset whose fingerprint closely matches a known well-conditioned 
regression benchmark may be expected to support penalized linear models, 
while one resembling a high-dimensional materials informatics dataset 
with strong spectral concentration may be better suited to kernel or 
graph-based methods. Extending this paradigm to physics-based modeling 
and simulation, the same proximity structure could inform the selection 
of constitutive model forms, numerical solver settings, or simulation 
initialization parameters based on the statistical characteristics of 
empirical calibration data, connecting the framework to validated 
modeling and simulation workflows in scientific computing. This positions 
the proposed framework as a foundation for data-driven model 
recommendation systems that leverage accumulated institutional knowledge 
about algorithm and simulation performance across heterogeneous scientific 
datasets.

    \section*{Acknowledgments}
    Work supported through the INL Laboratory Directed Research $\&$ Development (LDRD) Program under DOE Idaho Operations Office Contract DE-AC07-05ID14517.
    The authors of this work used Claude Sonnet 4.6 in the preparation of this document for general brainstorming and to improve grammar, sentence structure, and transitions. The authors have reviewed and take full responsibility for the resulting content.

\bibliography{references}

@article{agrawal2014exploration,
  title={Exploration of data science techniques to predict fatigue strength of steel from composition and processing parameters},
  author={Agrawal, Ankit and Deshpande, Parijat D and Cecen, Ahmet and Basavarsu, Gautham P and Choudhary, Alok N and Kalidindi, Surya R},
  journal={Integrating materials and manufacturing innovation},
  volume={3},
  number={1},
  pages={90--108},
  year={2014},
  publisher={Springer}
}

@article{dunn2020benchmarking,
  title={Benchmarking materials property prediction methods: the Matbench test set and Automatminer reference algorithm},
  author={Dunn, Alexander and Wang, Qi and Ganose, Alex and Dopp, Daniel and Jain, Anubhav},
  journal={npj Computational Materials},
  volume={6},
  number={1},
  pages={138},
  year={2020},
  publisher={Nature Publishing Group UK London}
}

@article{ward2016general,
  title={A general-purpose machine learning framework for predicting properties of inorganic materials},
  author={Ward, Logan and Agrawal, Ankit and Choudhary, Alok and Wolverton, Christopher},
  journal={npj Computational Materials},
  volume={2},
  number={1},
  pages={16028},
  year={2016},
  publisher={Nature Publishing Group}
}

@misc{de1988cohesion,
title = {Cohesion in metals. Transition metal alloys},
author = {De Boer, Frank R and Mattens, W and Boom, R and Miedema, AR and Niessen, AK},
volume = {1},
place = {Netherlands},
year = {1988},
month = {Jan}
}

@techreport{pham2025summary,
  title={Summary of Graphite Data Stored within NDMAS},
  author={Pham, Binh and Otani, Courtney and Rohrbaugh, David T},
  year={2025},
  institution={Idaho National Laboratory (INL), Idaho Falls, ID (United States)}
}

@misc{NDMAS_Graphite,
author       = {NDMAS},
title        = {Nuclear Data Management Analysis System: Graphite Materials Properties},
howpublished = {\url{https://ndmashome.inl.gov/htr/agcSite}},
note         = {Accessed: 2026-03-31}
}

@misc{citrination_mechanical_properties,
  author       = {Bajaj, Saurabh},
  title        = {Mechanical Properties of Some Steels},
  howpublished = {\url{https://citrination.com/datasets/153092/show_files/}},
  note         = {Accessed: 2026-03-31}
}

@misc{citrination_agrawal,
  author       = {Meredig, Bryce},
  title        = {Agrawal IMMI 2014},
  howpublished = {\url{https://citrination.com/datasets/150670/show_files/}},
  note         = {Accessed: 2026-03-31}
}

@article{heck1998corsika,
  title={CORSIKA: A Monte Carlo code to simulate extensive air showers},
  author={Heck, Dieter and Knapp, Johannes and Capdevielle, JN and Schatz, G and Thouw, T and others},
  journal={Report fzka},
  volume={6019},
  number={11},
  year={1998}
}

@misc{magic_gamma_telescope_159,
  author       = {Bock, R.},
  title        = {{MAGIC Gamma Telescope}},
  year         = {2004},
  howpublished = {UCI Machine Learning Repository},
  note         = {{DOI}: \url{https://doi.org/10.24432/C52C8B}}
}

@article{anderson1936species,
  title={The species problem in Iris},
  author={Anderson, Edgar},
  journal={Annals of the Missouri Botanical Garden},
  volume={23},
  number={3},
  pages={457--509},
  year={1936},
  publisher={JSTOR}
}

@article{fisher1936use,
  title={The use of multiple measurements in taxonomic problems},
  author={Fisher, Ronald A},
  journal={Annals of eugenics},
  volume={7},
  number={2},
  pages={179--188},
  year={1936},
  publisher={Wiley Online Library}
}

@misc{iris_53,
  author       = {Fisher, R. A.},
  title        = {{Iris}},
  year         = {1936},
  howpublished = {UCI Machine Learning Repository},
  note         = {{DOI}: \url{https://doi.org/10.24432/C56C76}}
}

@article{cortez2009modeling,
  title={Modeling wine preferences by data mining from physicochemical properties},
  author={Cortez, Paulo and Cerdeira, Ant{\'o}nio and Almeida, Fernando and Matos, Telmo and Reis, Jos{\'e}},
  journal={Decision support systems},
  volume={47},
  number={4},
  pages={547--553},
  year={2009},
  publisher={Elsevier}
}

@misc{wine_quality_186,
  author       = {Cortez, Paulo and Cerdeira, A. and Almeida, F. and Matos, T. and Reis, J.},
  title        = {{Wine Quality}},
  year         = {2009},
  howpublished = {UCI Machine Learning Repository},
  note         = {{DOI}: \url{https://doi.org/10.24432/C56S3T}}
}

@article{hollmann2025accurate,
  title={Accurate predictions on small data with a tabular foundation model},
  author={Hollmann, Noah and M{\"u}ller, Samuel and Purucker, Lennart and Krishnakumar, Arjun and K{\"o}rfer, Max and Hoo, Shi Bin and Schirrmeister, Robin Tibor and Hutter, Frank},
  journal={Nature},
  volume={637},
  number={8045},
  pages={319--326},
  year={2025},
  publisher={Nature Publishing Group UK London}
}

@article{qu2025tabicl,
  title={Tabicl: A tabular foundation model for in-context learning on large data},
  author={Qu, Jingang and Holzm{\"u}ller, David and Varoquaux, Ga{\"e}l and Morvan, Marine Le},
  journal={arXiv preprint arXiv:2502.05564},
  year={2025}
}

@article{van2024tabular,
  title={Why tabular foundation models should be a research priority},
  author={Van Breugel, Boris and Van Der Schaar, Mihaela},
  journal={arXiv preprint arXiv:2405.01147},
  year={2024}
}

@inproceedings{khanna2025tabular,
  title={Tabular embedding model (tem): Finetuning embedding models for tabular rag applications},
  author={Khanna, Sujit and Subedi, Shishir},
  booktitle={Intelligent Computing-Proceedings of the Computing Conference},
  pages={448--460},
  year={2025},
  organization={Springer}
}

@article{zhang2024vision,
  title={Vision-language models for vision tasks: A survey},
  author={Zhang, Jingyi and Huang, Jiaxing and Jin, Sheng and Lu, Shijian},
  journal={IEEE transactions on pattern analysis and machine intelligence},
  volume={46},
  number={8},
  pages={5625--5644},
  year={2024},
  publisher={IEEE}
}

@inproceedings{wu2019unified,
  title={Unified visual-semantic embeddings: Bridging vision and language with structured meaning representations},
  author={Wu, Hao and Mao, Jiayuan and Zhang, Yufeng and Jiang, Yuning and Li, Lei and Sun, Weiwei and Ma, Wei-Ying},
  booktitle={Proceedings of the IEEE/CVF Conference on Computer Vision and Pattern Recognition},
  pages={6609--6618},
  year={2019}
}

@article{li2025lost,
  title={Lost in embeddings: Information loss in vision-language models},
  author={Li, Wenyan and Tang, Raphael and Li, Chengzu and Zhang, Caiqi and Vulic, Ivan and S{\o}gaard, Anders},
  journal={arXiv preprint arXiv:2509.11986},
  volume={2},
  year={2025}
}

@article{dhillon2011multi,
  title={Multi-view learning of word embeddings via cca},
  author={Dhillon, Paramveer and Foster, Dean P and Ungar, Lyle},
  journal={Advances in neural information processing systems},
  volume={24},
  year={2011}
}

@article{fan2001variable,
  title={Variable selection via nonconcave penalized likelihood and its oracle properties},
  author={Fan, Jianqing and Li, Runze},
  journal={Journal of the American statistical Association},
  volume={96},
  number={456},
  pages={1348--1360},
  year={2001},
  publisher={Taylor \& Francis}
}

@article{gavish2014optimal,
  title={The optimal hard threshold for singular values is $4/\sqrt{3}$},
  author={Gavish, Matan and Donoho, David L},
  journal={IEEE Transactions on Information Theory},
  volume={60},
  number={8},
  pages={5040--5053},
  year={2014},
  publisher={IEEE}
}

@incollection{dwork2025differential,
  title={Differential privacy},
  author={Dwork, Cynthia},
  booktitle={Encyclopedia of Cryptography, Security and Privacy},
  year={2025}
}

@article{witten2009penalized,
  title={A penalized matrix decomposition, with applications to sparse principal components and canonical correlation analysis},
  author={Witten, Daniela M and Tibshirani, Robert and Hastie, Trevor},
  journal={Biostatistics},
  volume={10},
  number={3},
  pages={515--534},
  year={2009},
  publisher={Oxford University Press}
}

@article{wang2020minilm,
  title={Minilm: Deep self-attention distillation for task-agnostic compression of pre-trained transformers},
  author={Wang, Wenhui and Wei, Furu and Dong, Li and Bao, Hangbo and Yang, Nan and Zhou, Ming},
  journal={Advances in neural information processing systems},
  volume={33},
  pages={5776--5788},
  year={2020}
}

@article{vanschoren2018meta,
  title={Meta-learning: A survey},
  author={Vanschoren, Joaquin},
  journal={arXiv preprint arXiv:1810.03548},
  year={2018}
}

@article{feurer2015efficient,
  title={Efficient and robust automated machine learning},
  author={Feurer, Matthias and Klein, Aaron and Eggensperger, Katharina and Springenberg, Jost and Blum, Manuel and Hutter, Frank},
  journal={Advances in neural information processing systems},
  volume={28},
  year={2015}
}

@article{nargesian2019data,
  title={Data lake management: challenges and opportunities},
  author={Nargesian, Fatemeh and Zhu, Erkang and Miller, Ren{\'e}e J and Pu, Ken Q and Arocena, Patricia C},
  journal={Proceedings of the VLDB Endowment},
  volume={12},
  number={12},
  pages={1986--1989},
  year={2019},
  publisher={VLDB Endowment}
}

@article{rahm2001survey,
  title={A survey of approaches to automatic schema matching},
  author={Rahm, Erhard and Bernstein, Philip A},
  journal={the VLDB Journal},
  volume={10},
  number={4},
  pages={334--350},
  year={2001},
  publisher={Springer}
}

@inproceedings{zhu2019josie,
  title={Josie: Overlap set similarity search for finding joinable tables in data lakes},
  author={Zhu, Erkang and Deng, Dong and Nargesian, Fatemeh and Miller, Ren{\'e}e J},
  booktitle={Proceedings of the 2019 International Conference on Management of Data},
  pages={847--864},
  year={2019}
}

@article{chepurko2020arda,
  title={ARDA: automatic relational data augmentation for machine learning},
  author={Chepurko, Nadiia and Marcus, Ryan and Zgraggen, Emanuel and Fernandez, Raul Castro and Kraska, Tim and Karger, David},
  journal={arXiv preprint arXiv:2003.09758},
  year={2020}
}

@article{zha2025data,
  title={Data-centric artificial intelligence: A survey},
  author={Zha, Daochen and Bhat, Zaid Pervaiz and Lai, Kwei-Herng and Yang, Fan and Jiang, Zhimeng and Zhong, Shaochen and Hu, Xia},
  journal={ACM Computing Surveys},
  volume={57},
  number={5},
  pages={1--42},
  year={2025},
  publisher={ACM New York, NY}
}

@inproceedings{arik2021tabnet,
  title={Tabnet: Attentive interpretable tabular learning},
  author={Arik, Sercan {\"O} and Pfister, Tomas},
  booktitle={Proceedings of the AAAI conference on artificial intelligence},
  volume={35},
  number={8},
  pages={6679--6687},
  year={2021}
}

@article{somepalli2021saint,
  title={Saint: Improved neural networks for tabular data via row attention and contrastive pre-training},
  author={Somepalli, Gowthami and Goldblum, Micah and Schwarzschild, Avi and Bruss, C Bayan and Goldstein, Tom},
  journal={arXiv preprint arXiv:2106.01342},
  year={2021}
}

@article{lewis2020retrieval,
  title={Retrieval-augmented generation for knowledge-intensive nlp tasks},
  author={Lewis, Patrick and Perez, Ethan and Piktus, Aleksandra and Petroni, Fabio and Karpukhin, Vladimir and Goyal, Naman and K{\"u}ttler, Heinrich and Lewis, Mike and Yih, Wen-tau and Rockt{\"a}schel, Tim and others},
  journal={Advances in neural information processing systems},
  volume={33},
  pages={9459--9474},
  year={2020}
}

@inproceedings{reimers-2019-sentence-bert,
    title = "Sentence-BERT: Sentence Embeddings using Siamese BERT-Networks",
    author = "Reimers, Nils and Gurevych, Iryna",
    booktitle = "Proceedings of the 2019 Conference on Empirical Methods in Natural Language Processing",
    year = {2019}
}

@article{mazumder2010spectral,
  title={Spectral regularization algorithms for learning large incomplete matrices},
  author={Mazumder, Rahul and Hastie, Trevor and Tibshirani, Robert},
  journal={The Journal of Machine Learning Research},
  volume={11},
  pages={2287--2322},
  year={2010},
  publisher={JMLR. org}
}

@article{killick2012optimal,
  title={Optimal detection of changepoints with a linear computational cost},
  author={Killick, Rebecca and Fearnhead, Paul and Eckley, Idris A},
  journal={Journal of the American Statistical Association},
  volume={107},
  number={500},
  pages={1590--1598},
  year={2012},
  publisher={Taylor \& Francis}
}

@article{tibshirani1996regression,
  title={Regression shrinkage and selection via the lasso},
  author={Tibshirani, Robert},
  journal={Journal of the Royal Statistical Society Series B: Statistical Methodology},
  volume={58},
  number={1},
  pages={267--288},
  year={1996},
  publisher={Oxford University Press}
}

@Article{breheny2011coordinate,
  author = {Patrick Breheny and Jian Huang},
  title = {Coordinate descent algorithms for nonconvex penalized regression,
	with applications to biological feature selection},
  journal = {Annals of Applied Statistics},
  year = {2011},
  volume = {5},
  pages = {232--253},
  number = {1},
  doi = {10.1214/10-AOAS388},
  url = {https://doi.org/10.1214/10-AOAS388},
}

@article{killick2014changepoint,
  title={changepoint: An R package for changepoint analysis},
  author={Killick, Rebecca and Eckley, Idris A},
  journal={Journal of statistical software},
  volume={58},
  pages={1--19},
  year={2014}
}

@misc{kunz2026mungeR,
  author       = {Kunz, M. Ross},
  title        = {Modular Utilities for Numeric and General-purpose Embedding in R},
  year         = {2026},
  note         = {In preparation}
}
    \section*{Supplementary}\label{sec:supplementary}
    
\subsection{UCI Machine Learning Repository Datasets}
\label{subsec:supp:uci}

\subsubsection{Iris (iris)}
The Iris dataset contains 150 specimens across three species (\textit{Iris 
setosa}, \textit{Iris versicolor}, and \textit{Iris virginica}), with four 
continuous morphological measurements recorded per specimen~\citep{anderson1936species,
fisher1936use,iris_53}.

\begin{itemize}
    \item \textbf{Variables:} sepal.length, sepal.width, petal.length, 
    petal.width, class
\end{itemize}

\subsubsection{MAGIC Gamma Telescope (telescope)}
This dataset contains observations generated via Monte Carlo simulation of 
gamma-ray and hadron shower events recorded by an imaging atmospheric 
Cherenkov telescope~\citep{heck1998corsika,magic_gamma_telescope_159}. 
Continuous features describe the geometric and photometric properties of 
each shower image, and the binary response distinguishes signal (gamma) 
from background (hadron) events.

\begin{itemize}
    \item \textbf{Variables:} fLength, fWidth, fSize, fConc, fConc1, fAsym, 
    fM3Long, fM3Trans, fAlpha, fDist, class
\end{itemize}

\subsubsection{Wine Quality (wine red and wine white)}
The Wine Quality datasets consist of physicochemical measurements for red 
and white Vinho Verde wine samples from northern Portugal~\citep{cortez2009modeling,
wine_quality_186}. The response is an ordinal quality score assigned by 
human sensory evaluation, commonly treated as either a regression or 
discretized classification target. The red and white subsets are treated 
as independent datasets in this work due to differences in composition 
profiles and class distributions.

\begin{itemize}
    \item \textbf{Variables:} fixed.acidity, volatile.acidity, citric.acid, 
    residual.sugar, chlorides, free.sulfur.dioxide, total.sulfur.dioxide, 
    density, pH, sulphates, alcohol, quality
\end{itemize}

\begin{table}[htbp]
    \centering
    \caption{Dimensions of UCI Machine Learning Repository datasets.}
    \label{tab:general_datasets}
    \begin{tabular}{p{0.4\textwidth} | p{0.25\textwidth} p{0.25\textwidth}}
        Data Set & Num Rows & Num Cols \\
        \hline
        iris          &    150 &  5 \\
        telescope     &  19020 & 11 \\
        wine\_red     &   1599 & 12 \\
        wine\_white   &   4898 & 12 \\
    \end{tabular}
\end{table}

\subsection{Steel and Materials Property Datasets}
\label{subsec:supp:steels}

\subsubsection{Citrine Agrawal Steel Strength (citrine agrawal)}
This dataset, hosted on the Citrination platform and introduced by 
\cite{agrawal2014exploration,citrination_agrawal}, was 
originally assembled from the National Institute for Materials Science 
(NIMS) public-domain database to support data-driven prediction of steel 
fatigue strength. Compositional and processing parameters serve as input 
features, and fatigue strength (MPa) is the regression target. Features 
span elemental composition fractions and thermo-mechanical processing 
conditions including reduction ratio, quenching temperature, and 
tempering temperature.

\begin{itemize}
    \item \textbf{Variables:} Normalizing Temperature C, Through Hardening 
    Temperature C, Through Hardening Time min, Cooling Rate for Through 
    Hardening C hr, Carburization Temperature C, Carburization Time min, 
    Diffusion Temperature C, Diffusion Time min, Quenching Media Temperature 
    for Carburization C, Tempering Temperature C, Tempering Time min, Cooling 
    Rate for Tempering C hr, Reduction Ratio Ingot to Bar, Area Proportion of 
    Inclusions Deformed by Plastic Work, Area Proportion of Inclusions 
    Occurring in Discontinuous Array, Area Proportion of Isolated Inclusions, 
    Fatigue Strength MPa, C at pct, Cr at pct, Cu at pct, Fe at pct, Mn at 
    pct, Mo at pct, Ni at pct, P at pct, S at pct, Si at pct
\end{itemize}

\subsubsection{Citrine Steels (citrine steel)}
This dataset contains mechanical property measurements for a collection of 
steel alloys described by elemental composition \citep{citrination_mechanical_properties}. Feature representations 
are derived directly from alloy composition without additional processing 
parameters, and measured properties include yield strength, ultimate tensile 
strength, elongation, Charpy impact energy, and fracture toughness.

\begin{itemize}
    \item \textbf{Variables:} Yield strength YS MPa, Yield strength YS MPa 
    temperature C, Ultimate tensile strength UTS MPa, Ultimate tensile 
    strength UTS MPa temperature C, Elongation, Elongation temperature C, 
    Charpy impact energy J, Charpy impact energy J temperature C, Fracture 
    toughness K\textsubscript{IC} MPa m, Fracture toughness 
    K\textsubscript{IC} MPa m temperature C, Al wt pct, C wt pct, Co wt pct, 
    Cr wt pct, Mn wt pct, Mo wt pct, N wt pct, Nb wt pct, Ni wt pct, Si wt 
    pct, Ti wt pct, V wt pct
\end{itemize}

\subsubsection{Matbench Steels (matbench steels)}
The Matbench steels task is part of the MatBench v0.1 benchmark suite 
developed by the Materials Project for standardized evaluation of machine 
learning models on materials property prediction~\citep{dunn2020benchmarking}. 
The dataset contains 312 steel alloy compositions with yield strength (MPa) 
as the regression target, with features constructed from elemental 
composition strings spanning the full periodic table.

\begin{itemize}
    \item \textbf{Variables:} yield.strength, H, He, Li, Be, B, C, N, O, F, 
    Ne, Na, Mg, Al, Si, P, S, Cl, Ar, K, Ca, Sc, Ti, V, Cr, Mn, Fe, Co, Ni, 
    Cu, Zn, Ga, Ge, As, Se, Br, Kr, Rb, Sr, Y, Zr, Nb, Mo, Tc, Ru, Rh, Pd, 
    Ag, Cd, In, Sn, Sb, Te, I, Xe, Cs, Ba, La, Ce, Pr, Nd, Pm, Sm, Eu, Gd, 
    Tb, Dy, Ho, Er, Tm, Yb, Lu, Hf, Ta, W, Re, Os, Ir, Pt, Au, Hg, Tl, Pb, 
    Bi, Po, At, Rn, Fr, Ra, Ac, Th, Pa, U, Np, Pu, Am, Cm, Bk, Cf, Es, Fm, 
    Md, No, Lr, Rf, Db, Sg, Bh, Hs, Mt, Ds, Rg, Cn, Nh, Fl, Mc, Lv, Ts, Og
\end{itemize}

\subsubsection{MAGPIE (magpie)}
The MAGPIE (Materials Agnostic Platform for Informatics and Exploration) 
feature set was introduced by \cite{ward2016general} as a 
general-purpose descriptor framework for inorganic materials. MAGPIE defines 
a composition-based featurization scheme that computes statistics (mean, 
range, minimum, maximum, and mean absolute deviation) over 22 elemental 
properties for any stoichiometric formula, yielding a fixed-length numerical 
descriptor. In this work, MAGPIE features serve as the input representation 
for steel compositions drawn from the datasets described above, providing a 
physics-informed alternative to raw composition fractions.

\begin{itemize}
    \item \textbf{Variables:} MeltingT, Miracle Radius, Covalent Radius, 
    First Ionization Energy, GSest FCClatcnt, Electronegativity, Density, 
    ZungerPP.r s, Dipole Polarizability, GSest BCClatcnt, Log Thermal 
    Conductivity, NsValence, GSmagmom, Polarizability, NValence, IsAlkali, 
    AtomicVolume, phi, Number, IsNonmetal, NfValence, HHIp, VdWRadius, HHIr, 
    Np Unfilled, Bulk Modulus, ZungerPP.r p, NsUnfilled, ZungerPP.r sigma, 
    Thermal Conductivity, Atomic Radius, Shear Modulus, Is Metalloid, 
    NfUnfilled, Second Ionization Energy, IsDBlock, Atomic Weight, Heat 
    Vaporization, Mendeleev Number, GSvolume pa, NU nfilled, IsMetal, 
    Electron Affinity, NdUnfilled, NdValence, MolarVolume, Column, 
    ICSDVolume, IsFBlock, Space GroupNumber, Heat Capacity Mass, Allen 
    Electronegativity, BoilingT, GSenergy pa, HeatCapacity Molar, n ws.third, 
    HeatFusion, ZungerPP.r d, ZungerPP.r pi, NpValence, Row, Fusion Enthalpy, 
    GSbandgap
\end{itemize}

\subsubsection{Miedema (miedema)}
The Miedema dataset encodes binary alloy formation enthalpy estimates derived 
from the semi-empirical Miedema model~\citep{de1988cohesion}, which 
characterizes intermetallic interactions through elemental parameters 
including electron density at the Wigner-Seitz cell boundary and 
electronegativity-related work function differences. These pairwise elemental 
descriptors provide thermodynamic context for alloy stability and are 
incorporated as supplementary features in the materials property prediction 
tasks examined here.

\begin{itemize}
    \item \textbf{Variables:} molar volume, electron density, 
    electronegativity, valence electrons, a const, R const, H trans, 
    compressibility, shear modulus, melting point, structural stability
\end{itemize}

\begin{table}[htbp]
    \centering
    \caption{Dimensions of materials informatics datasets.}
    \label{tab:steels_datasets}
    \begin{tabular}{p{0.4\textwidth} | p{0.25\textwidth} p{0.25\textwidth}}
        Data Set & Num Rows & Num Cols \\
        \hline
        citrine\_agrawal &  437 &  27 \\
        citrine\_steel   &  842 &  22 \\
        matbench\_steel  &  312 & 119 \\
        magpie           &  118 &  63 \\
        miedema          &   73 &  11 \\
    \end{tabular}
\end{table}

\subsection{NDMAS Graphite Datasets}
\label{subsec:supp:graphite}

\subsubsection{Bulk Density (BulkDensity)}
Bulk density records in NDMAS are derived from dimensional and mass 
measurements collected as part of the baseline graphite physical properties 
characterization program. Specimen bulk density is computed from measured 
mass, diameter, and length, with geometry conforming to ASTM standard 
cylindrical specimen configurations.

\begin{itemize}
    \item \textbf{Variables:} amb temp, amb humidity, specimen length avg, 
    specimen diam avg, mass, bulk density
\end{itemize}

\subsubsection{Compression (Compression)}
Compressive strength measurements are obtained from cylindrical specimens 
machined to ASTM test standard geometries, capturing load, stress, 
displacement, and strain at maximum load.

\begin{itemize}
    \item \textbf{Variables:} amb temp, amb humidity, stress compr at max 
    load, load compr at max load, displacement at max load, strain compr
\end{itemize}

\subsubsection{Flexure (Flex)}
Flexural strength measurements are collected from specimens tested in 
three-point or four-point bending configurations per ASTM standard 
geometries, recording maximum load, maximum flexure stress, mid-span 
deflection, and elapsed time at peak load.

\begin{itemize}
    \item \textbf{Variables:} amb temp, amb humidity, pmaxflex, stress flex 
    max, deflect mid at pmax, elapsed time at pmax
\end{itemize}

\subsubsection{Gas Chemistry (GasChem)}
Records of gas composition captured during AGC irradiation campaigns 
characterize the helium purge environment surrounding irradiation capsules. 
These data serve as irradiation condition monitoring records rather than 
material property measurements.

\begin{itemize}
    \item \textbf{Variables:} CO2 Conc, CO Conc, Ar Conc
\end{itemize}

\subsubsection{Irradiation Summary (IRRSummary)}
Summary records of irradiation conditions for AGC specimens include fast 
neutron fluence, irradiation temperature, and applied stress. These data 
provide the experimental context for interpreting post-irradiation property 
changes and are linked to the characterization datasets by specimen 
identifier.

\begin{itemize}
    \item \textbf{Variables:} AvgTemperature, CumeDoseDPA, PowerWgtedLoad
\end{itemize}

\subsubsection{Resistivity (Resistivity)}
Electrical resistivity measurements are collected on graphite specimens 
prior to mechanical testing using standardized four-point contact methods 
per ASTM C611-05, providing a non-destructive characterization of the 
electrical transport properties of each specimen. Resistivity values are 
tightly distributed across specimens, reflecting the relatively uniform 
microstructural character of the nuclear-grade graphite grades tested.

\begin{itemize}
    \item \textbf{Variables:} amb temp, amb humidity, resistance, 
    resistivity, potentialmeasure, AppliedCurrent, ComplianceVoltage
\end{itemize}

\subsubsection{Resonance (Resonance)}
Resonance measurements characterize the elastic constants of nuclear-grade 
graphite specimens through fundamental frequency analysis in both flexural 
and torsional vibration modes, yielding dynamic Young's modulus and shear 
modulus per ASTM C747-93 and ASTM C1259-08. The parallelepiped geometry 
of the flexural specimens is particularly well-suited for this measurement, 
providing accurate elastic constant determination from a single 
non-destructive test.

\begin{itemize}
    \item \textbf{Variables:} amb temp, amb humidity, resonance freq avg, 
    elastic modulus flex
\end{itemize}

\subsubsection{Sonic Velocity (SonicVelocity)}
Measurements of ultrasonic pulse velocity through graphite specimens serve 
as a non-destructive indicator of material condition and are used to derive 
elastic moduli including longitudinal, shear, and corrected elastic modulus 
alongside Poisson's ratio.

\begin{itemize}
    \item \textbf{Variables:} vel sonic shear, elastic modulus long, elastic 
    modulus shear, elastic modulus corrected, amb temp, amb humidity, 
    poissons ratio
\end{itemize}

\subsubsection{Specimen Neutronics (SpecNeut)}
Computed physics records for AGC specimens include fast neutron fluence and 
displacements per atom (dpa) derived from reactor physics calculations. 
These records quantify the radiation dose received by each specimen and 
serve as primary input variables for property change modeling.

\begin{itemize}
    \item \textbf{Variables:} Fluence, DPA
\end{itemize}

\subsubsection{Specimen Thermal (SpecThermal)}
Specimen-level temperature records are derived from thermocouple 
measurements and thermal analysis of AGC capsule configurations, 
characterizing the thermal environment experienced by individual specimens 
during irradiation.

\begin{itemize}
    \item \textbf{Variables:} Temperature, ElevAboveMidCore
\end{itemize}

\subsubsection{Tensile (Tensile)}
Tensile testing is performed on cylindrical dog-bone specimens per ASTM 
C749-08, capturing ultimate tensile strength, stress at break, strain at 
break, and elastic modulus for each specimen extracted from the graphite 
billet. Gauge diameters are verified against ASTM dimensional tolerances 
prior to testing, and extensometer-based strain measurements provide 
additional correlation with stress values through material elastic constants.

\begin{itemize}
    \item \textbf{Variables:} amb temp, amb humidity, stress rate, elastic 
    modulus tensile, ultimate tensile strength, p at break, stress at break, 
    strain1 at break, strain2 at break, strainavg at break
\end{itemize}

\subsubsection{Thermal Conductivity (ThermalConductivity)}
Thermal diffusivity and thermal conductivity measurements are obtained from 
laser flash analysis on disk-shaped specimens per ASTM E1461-07. Thermal 
conductivity is particularly sensitive to radiation-induced lattice damage 
and is a primary design parameter for HTR core thermal analysis.

\begin{itemize}
    \item \textbf{Variables:} Temperature C, Thermal Conductivity W per mK
\end{itemize}

\begin{table}[htbp]
    \centering
    \caption{Dimensions of NDMAS graphite datasets.}\label{tab:ndmas_datasets}
    \begin{tabular}{p{0.4\textwidth} | p{0.25\textwidth} p{0.25\textwidth}}
        Data Set & Num Rows & Num Cols \\
        \hline
        BulkDensity         & 14949 &  6 \\
        Compression         &  2206 &  6 \\
        Flex                &  2351 &  6 \\
        GasChem             &   192 &  3 \\
        IRRSummary          &  1311 &  3 \\
        Resistivity         &  2733 &  7 \\
        Resonance           &  7311 &  4 \\
        SonicVelocity       &  4449 &  7 \\
        SpecNeut            & 12691 &  2 \\
        SpecThermal         & 26912 &  2 \\
        Tensile             &  2296 & 10 \\
        ThermalConductivity &  3729 &  2 \\
    \end{tabular}
\end{table}

\subsection{Statistical Descriptors List}

\begin{table}[htbp]
\centering
\caption{Statistical descriptors computed at the multivariate and univariate level.}
\label{tab:munge_descriptors}
\begin{tabular}{p{0.45\textwidth} p{0.45\textwidth}}
\toprule
\textbf{Multivariate Descriptors} & \textbf{Univariate Descriptors} \\
\midrule

\textit{Counts} & \textit{Bounds} \\
\hspace{1em} Total count & \hspace{1em} Minimum \\
\hspace{1em} Total unique & \hspace{1em} Maximum \\
\hspace{1em} Missing count & \hspace{1em} Range \\
\hspace{1em} Cardinality percent & \hspace{1em} Negative count \\
\hspace{1em} Non numeric percent & \hspace{1em} \\[4pt]

\textit{Norms} & \textit{Norms} \\
\hspace{1em} $\ell_1$ norm & \hspace{1em} $\ell_0$ norm \\
\hspace{1em} Frobenius norm & \hspace{1em} $\ell_1$ norm \\
\hspace{1em} Infinity norm & \hspace{1em} $\ell_2$ norm \\
\hspace{1em} Maximum Modulus & \\[4pt]

\textit{Singular Values} & \textit{Moments \& Quantiles} \\
\hspace{1em} Numerical rank & \hspace{1em} Mean \\
\hspace{1em} Spectral norm & \hspace{1em} Standard deviation \\
\hspace{1em} Condition number & \hspace{1em} Skewness \\
\hspace{1em} Frobenius norm & \hspace{1em} Kurtosis \\
\hspace{1em} Nuclear norm & \hspace{1em} Coefficient of Variation \\
\hspace{1em} Entropy & \hspace{1em}  Q1 \\
\hspace{1em} Median gap & \hspace{1em}  Median \\
\hspace{1em} Maximum gap & \hspace{1em}  Q3 \\[4pt]

\textit{Univariate Correlations} &  \textit{Univariate Least Squares} \\
\hspace{1em} Pairwise correlation & \hspace{1em} Autoregressive 1-lag \\
\hspace{1em} Pairwise regression coefficients & \hspace{1em} Slope via index \\[4pt]

\textit{Multivariate Correlations} & \textit{Autocorrelation} \\
\hspace{1em} SCAD multivariate correlation &  \hspace{1em} Autocorrelation significant lags \\
\hspace{1em} Multivariate regression coefficients & \hspace{1em} Partial autocorrelation significant lags \\[4pt]

& \textit{Frequency (FFT)} \\
& \hspace{1em} Dominant frequency \\
& \hspace{1em} Phase information \\[4pt]

& \textit{Change Point Detection} \\
& \hspace{1em} Mean shift locations \\
& \hspace{1em} Variance shift locations \\
& \hspace{1em} Mean-variance shift locations \\[4pt]

\bottomrule
\end{tabular}
\end{table}
\end{document}